\documentclass{article}

\PassOptionsToPackage{numbers, compress}{natbib}
\usepackage[final]{nips_2016}

\usepackage[utf8]{inputenc} 
\usepackage[T1]{fontenc}    
\usepackage{url}            
\usepackage{booktabs}       
\usepackage{amsfonts}       
\usepackage{nicefrac}       
\usepackage{microtype}      
\usepackage{url}
\usepackage{graphicx}
\usepackage{amsfonts}
\usepackage{amsmath}
\usepackage{times}
\usepackage{fancyhdr}
\usepackage{algo}
\usepackage{wrapfig}
\usepackage{caption}
\usepackage{subcaption}
\usepackage{epsfig}
\usepackage{epstopdf}
\usepackage{algorithm}
\usepackage{algpseudocode}
\usepackage{amsmath,bm}
\usepackage{color}
\usepackage{authblk}
\usepackage{mathtools}

\def\Fb{\mathbf{F}}
\def\Ib{\mathbf{I}}

\def\qb{\mathbf{q}}

\def\vb{\mathbf{v}}\def\wb{\mathbf{w}}

\def\Qbb{\mathbb{Q}}

\def\Q{\Qbb}

\newcommand{\defeq}{\vcentcolon=}

\newcommand{\E}{\mathrm{E}}
\newcommand{\Var}{\mathrm{Var}}
\newcommand{\Cov}{\mathrm{Cov}}

\newcommand{\es}{{\text{es}}}
\newcommand{\ec}{{\text{ec}}}

\title{Adaptive Probabilistic Trajectory Optimization via Efficient Approximate Inference}

\author[Yunpeng Pan et al.]
       {\textbf{Yunpeng Pan$^{1,2}$, Xinyan Yan$^{1,3}$, Evangelos Theodorou$^{1,2}$, and Byron Boots$^{1,3}$}\\
       $^1$Institute for Robotics and Intelligent Machines, Georgia Institute of Technology\\
       $^2$School of Aerospace Engineering, Georgia Institute of Technology \\ $^3$School of Interactive Computing, Georgia Institute of Technology \\
       \{ypan37,xyan43,evangelos.theodorou\}@gatech.edu, bboots@cc.gatech.edu\\ 
       }

\newcommand{\rd}{{\mathrm d}}
\newcommand{\rp}{{\mathrm p}}
\newcommand{\rtr}{{\mathrm{tr}}}
\newcommand{\rT}{{\mathrm{T}}}
\newcommand{\va}{{\bf a}}
\newcommand{\vc}{{\bf c}}
\newcommand{\vx}{{\bf x}}

\newcommand{\vq}{{\bf q}}

\newcommand{\vs}{{\bf s}}
\newcommand{\vf}{{\bf f}}

\newcommand{\ve}{{\bf e}}

\newcommand{\vu}{{\bf u}}
\newcommand{\vv}{{\bf v}}
\newcommand{\vI}{{\bf I}}

\newcommand{\vl}{{\bf l}}

\newcommand{\vw}{{\bf w}}

\newcommand{\vR}{{\bf R}}

\newcommand{\vL}{{\bf L}}

\newcommand{\vF}{{\bf F}}

\newcommand{\vD}{{\bf D}}

\newcommand{\vV}{{\bf V}}

\newcommand{\vT}{{\bf T}}
\newcommand{\vC}{{\bf C}}

\newcommand{\vQ}{{\bf Q}}

\newcommand{\vS}{{\bf S}}
\newcommand{\vX}{{\bf X}}

\newcommand{\vP}{{\bf P}}
\newcommand{\vA}{{\bf A}}

\newcommand{\tx}{{\tilde{\bf x}}}

\newcommand{\tX}{{\tilde{\bf X}}}

\newcommand{\VPhi}{{\mbox{\boldmath$\Phi$}}}
\newcommand{\vPhi}{{\mbox{\boldmath$\Phi$}}}
\newcommand{\VOmega}{{\mbox{\boldmath$\Omega$}}}

\newcommand{\vphi}{{\mbox{\boldmath$\phi$}}}

\newcommand{\VSigma}{{\mbox{\boldmath$\Sigma$}}}

\newcommand{\vmu}{{\bm{\mu}}}
\newcommand{\vSigma}{{\mbox{\boldmath$\Sigma$}}}

\newcommand{\vomega}{{\bm{\omega}}}

\newcommand{\vtheta}{{\mbox{\boldmath$\theta$}}}

\newcommand{\T}{^\mathsf{T}}

\newcommand{\mF}{\mathcal{F}}
\newcommand{\tmu}{\tilde{\bm{\mu}}}

\newcommand{\tSigma}{{\tilde{\bf \vSigma}}}

\DeclareMathOperator{\Tr}{Tr}
\begin{document}

\maketitle
\vspace{-1mm}
\begin{abstract}\vspace{-1mm}
Robotic systems must be able to quickly and robustly make decisions when operating in uncertain and dynamic environments. While Reinforcement Learning (RL) can be used to compute optimal policies with little prior knowledge about the environment, it suffers from slow convergence. 
An alternative approach is Model Predictive Control (MPC), which optimizes policies quickly, 
but also requires accurate models of the system dynamics and environment.
In this paper we propose a new approach, adaptive probabilistic trajectory optimization, that combines the benefits of RL and MPC. Our method uses scalable approximate inference to learn and updates probabilistic models in an online incremental fashion while also computing optimal control policies via successive local approximations. 
We present two variations of our algorithm 
based on the Sparse Spectrum Gaussian Process (SSGP) model, 
and we test our algorithm on three learning tasks, demonstrating the effectiveness and efficiency of our approach.
\end{abstract}
\vspace{-0.3 cm}
\section{Introduction}
\vspace{-0.1 cm}
Over the last decade, reinforcement learning (RL) has started to be successfully applied to robotics and autonomous systems.  While model-free RL has demonstrated promising results~\cite{deisenroth2013survey}, it typically requires human expert demonstrations or relies on lots of direct interactions with physical systems. Model-based RL was developed to address the issue of sample inefficiency by learning dynamics models explicitly from  data, which can provide better generalization~\cite{atkeson1997comparison,deisenroth2013survey}. However, these deterministic model-based methods suffer from error in the learned models which compounds when making long-range predictions. 
Recent probabilistic model-based RL methods overcome this issue, achieving  state-of-the-art performance~\cite{deisenroth2014gaussian,pan2014probabilistic,kupcsik2014model}. These methods represent dynamics models using nonparametric Gaussian processes (GPs) and take into account model uncertainty for control policy learning. Despite these successes, inference in nonparametric probabilistic models can be computationally demanding, making it difficult to adapt to rapid changes in the environment or dynamics. 

In contrast to probabilistic model-based RL methods, where learning  through  interaction occurs at the scale of trajectories/rollouts, Model Predictive Control (MPC) is more reactive, performing re-optimization every few timesteps. However, MPC also requires accurate models of the system dynamics and environment. Our goal in this work is to develop a method that combines the benefits of MPC  with  probabilistic model-based RL to perform learning control at small time scales with little prior knowledge of the dynamics. 
More precisely, we propose a method that relies on local trajectory optimization such as DDP \cite{jacobson1970differential},  which is  scalable and has been shown to work well on challenging learning control tasks in robotics~\cite{abbeel2007application,tassacontrol,levine2014learning}. 
We model the dynamics probabilistically 
and perform optimization in belief space.   
In previous GP-related RL methods, approximate inference is the major computational bottleneck ~\cite{deisenroth2014gaussian,pan2014probabilistic,kupcsik2014model}. Therefore, we develop two novel scalable approximate inference algorithms based on Sparse Spectrum Gaussian Processes (SSGPs)~\cite{Lazaroetal10}.  To cope with changes in the task or with varying dynamics  we apply online re-optimization in the spirit of MPC.  
  This combination leads to a general data-driven framework for online trajectory optimization.

%
%

\vspace{-0.2 cm}
\section{Trajectory Optimization}
\vspace{-0.2 cm}
We consider a general unknown dynamical system described by the following differential equation
\begin{equation}\label{real_dyn}
      \rd \vx = \Fb(\vx,\vu)  \rd t + \vC \rd{\bf \omega},~~ \quad \vx(t_{0})  = \vx_{0}, ~~\quad \rd{\bf \omega}\sim \mathcal{N}(0,\vSigma_{\omega}),
 \end{equation}
where $\vx\in\mathbb{R}^n$ is the state,  $\vu\in\mathbb{R}^m$ is the control   and  ${\bf\omega}\in\mathbb{R}^p$ is standard Brownian noise.  The goal of optimal control and reinforcement learning is to find the control policy $\pi(\vx(t),t)$ that minimizes the expected cost \vspace{-3mm}
\begin{equation}\label{expected_cost}
J^{\pi}(\vx(t_{0}))= \E_{\vx}\Big[ h\big(\vx(T)\big)+\int^{T}_{t_{0}}\mathcal{L}\big(\vx(t),\pi(\vx(t)),t \big) \rd t \Big],
\end{equation} 
where $h(\vx(T))$ is the terminal cost, and $\mathcal{L}(\vx(t),\pi(\vx(t)),t)$ is the instantaneous cost rate. The control policy  $\vu(t) = \pi(\vx(t),t)$ is a function that maps states and time to controls. The  cost  $J^{\pi}(\vx(t_0))$ is defined as the expectation of the total cost accumulated from $t_{0}$ to $T$. $\E_{\vx}$ denotes the expectation operator with respect to $\vx$. We assume that the states are fully observable.  For the rest of our analysis, we discretize the time as $k=1,2,...,H$ with time step $\Delta t=\frac{T}{H-1}$ and denote $  \vx_{k} = \vx(t_k)$. We use this subscript rule for other time-varying variables as well. The discretized system dynamics can be written as $\vx_{k+1}=\vx_k+\Delta\vx_k$ where $\Delta\vx_k=\Delta t\Fb(\vx_k,\vu_k)$.  To simplify notation we define $\vf(\vx_k,\vu_k)=\Delta t \vF(\vx_k,\vu_k)$.
Throughout the paper we consider the quadratic instantaneous cost function
$
\mathcal{L} (\vx_k,\vu_k) = (\vx_k-\vx_k^{goal})^{\rT}\vQ(\vx_k-\vx_k^{goal}) + \vu_k^{\rT}\vR\vu_k,
$
where $\vQ$ and $\vR$ are weighting matrices. The problem formulation (\ref{expected_cost}) is a standard finite horizon control or RL problem. However, in this paper we will perform this optimization at every time step in a receding horizon fashion.

\paragraph{Differential Dynamic Programming (DDP)} DDP is a model-based trajectory optimization method for solving  optimal control problems defined in (\ref{expected_cost}). The main idea is that a complex nonlinear control problem can be simplified using local approximations such that the original problem becomes a linear-quadratic problem in the neighborhood of a trajectory.   In DDP, and related methods such as iLQG~\cite{todorov2005generalized}, a local model is constructed based on i) a first or second-order linear approximation of the  dynamics model; ii) a second-order local approximation of the  value function along a nominal trajectory. The optimal control law and value function can be computed in a backward pass. The control law is used to generate a new state-control trajectory in a forward pass, this trajectory  becomes the new nominal trajectory for the next iteration. DDP uses the aforementioned backward-forward pass to optimize the trajectory iteratively until convergence to an optimal solution. DDP-related methods have been widely used for solving control problem in robotics tasks~\cite{abbeel2007application,tassacontrol}.

\vspace{-0.1 cm}
\section{Probabilistic Model Learning and Inference}
\vspace{-0.1 cm}
DDP requires an accurate dynamics model, however a good analytic model is not always available. Therefore, various methods have been used within trajectory optimization frameworks to \emph{learn} a model. In iLQG-LD \cite{mitrovic2010adaptive} and Minimax DDP\cite{morimoto2002minimax}, the dynamics are learned using Locally Weighted Projection Regression (LWPR) \cite{vijayakumar2005incremental} and Receptive Field Weighted Regression (RFWR) \cite{schaal1998constructive}. However, both methods require a large amount of training data collected from lots of interactions with the physical system. Recently, probabilistic methods such as PDDP \cite{pan2014probabilistic} and AGP-iLQR \cite{boedecker2014approximate} have demonstrated impressive data efficiency with Gaussian process (GP) dynamics models. 
However, GP inference in PDDP \cite{pan2014probabilistic} is too computationally expensive for online optimization and incremental updates. AGP-iLQR \cite{boedecker2014approximate} employs subset of regression (SOR) approximations that feature faster GP inference and incremental model adaptation, but neglects the predictive uncertainty when performing multi-step predictions using the learned forward dynamics. This leads to biased long-range predictions. In order to perform efficient and robust trajectory optimization, we introduce a learning and inference scheme based on Sparse Spectrum Gaussian Processes (SSGPs)~\cite{Lazaroetal10,gijsberts2013real}.

\vspace{-0.1 cm}
\subsection{Model learning via sparse spectrum Gaussian processes}\label{SSGP}
\vspace{-0.1 cm}

Learning a continuous mapping from state-control pairs $\tx=(\vx,\vu)\in\mathbb{R}^{n+m}$ to state transitions $\Delta\vx\in\mathbb{R}^n$ can be viewed as probabilistic inference. 
Given a sequence of $N$ state-control pairs 
$\tX = \{(\vx_{i},\vu_i)\}_{i=1}^N$
and the corresponding state transition
$\Delta\vX = \{\Delta\vx_i\}_{i=1}^N$, Gaussian processes (GP) can be leveraged to learn the dynamics model \cite{deisenroth2014gaussian,pan2014probabilistic,kupcsik2014model}. 
The posterior distribution of the state transition at a test state-control pair can be computed in close-form by conditioning on the observations, because they are jointly Gaussian.
Although GP regression is a powerful regression technique, it exhibits significant practical limitations for learning and inference on large datasets due to its $O(N^3)$ computation and $O(N^2)$ space complexity, which is a direct consequence of having to store and invert a $N\times N$ matrix. This computational inefficiency is a bottleneck for applying GP-based RL in real-time.

\paragraph{Sparse Spectrum GP Regression (SSGPR)} SSGP~\cite{Lazaroetal10} is a recent approach that provides a principled approximation of GPR by employing a random Fourier feature approximation of the kernel function~\cite{rahimi2007random}. Based on \textit{Bochner's theorem}~\cite{Rudin62}, any shift-invariant kernel functions can be represented as the Fourier transform of a unique measure %
$ k(\tx_i - \tx_j) = \int_{\mathbb{R}^{n+m}} e^{i\vomega\T (\tx_i - \tx_j)}p(\vomega) d \vomega = \E_{\vomega} [\vphi_{\vomega}(\tx_i)\T \vphi_{\vomega}(\tx_j)]$. %
We can, therefore, unbiasedly approximate any shift-invariant function by drawing $r$ random samples from the distribution $p(\vomega)$, 
$k(\tx_i, \tx_j) \approx \sum\nolimits_{i=1}^r \vphi_{\vomega_i}(\tx_i)\T \vphi_{\vomega_i}(\tx_j)= \vphi(\tx_i) \T \vphi(\tx_j)$, 
where $\vphi(\tx)$ is a \textit{feature mapping} which maps a state-control pair to feature space.

We consider the popular Squared Exponential (SE) covariance function with Automatic Relevance Determination (ARD) distance measure as it has been applied successfully in learning dynamics and optimal control~\cite{gijsberts2013real,boedecker2014approximate},
$k(\tx_i,\tx_j) = \sigma_f^2 \exp ( -\frac{1}{2} (\tx_i - \tx_j)\T \vP^{-1} (\tx_i - \tx_j) )$,
where 
$\vP = \text{diag}(\{l^2_i\}_{i=1}^{n+m})$. 
The hyper-parameters consist of the signal variance $\sigma_f^2$, the noise variance $\sigma_n^2$ and the length scales $\vl = \{l_i\}_{i=1}^{n+m}$.
The feature mapping for this SE kernel can be derived from
$ \quad \vphi_{\vomega}(\tx) = \frac{\sigma_f}{\sqrt{r}} 
 [\begin{array}{cc} 
 \cos(\vomega\T\tx) &
 \sin(\vomega\T\tx)
 \end{array}] \T
$, and $\vomega \sim \mathcal{N}(0, \vP^{-1})$.
Assuming the prior distribution of weights of the features  $\vw \sim \mathcal{N}(0, \VSigma_p)$, the posterior distribution of $\Delta \vx$ can be derived as in standard Bayesian linear regression
\begin{equation}\label{pred_mean}
\Delta\vx | \tX, \Delta \vX, \tx \sim \mathcal{N} (
\vw\T \vphi, \sigma_n^2 (1+ \vphi\T \vA^{-1}\vphi)),
\end{equation}
where
$
\vw = \vA^{-1}\VPhi \Delta \vX$,
$\vA = \VPhi \VPhi\T + \sigma_n^2\VSigma_p^{-1}$,
$\vphi = \vphi(\tx)$,
$\VPhi = \{\vphi(\tx_i)\}_{i=1}^{r}$\normalsize
. Thus the computational complexity becomes $O(Nr^2+r^3)$, which is significantly more efficient than GPR with $O(N^3)$ time complexity when the number of random features $r$ is much smaller than the number of training samples $N$.  The hyper-parameters can be learned by maximizing the log-likelihood of the training outputs given the inputs using numerical methods such as conjugate gradient \cite{williams2006gaussian}.

\paragraph{On-line model adaptation}
To update the weights $\vw$ incrementally given a new sample, we do not store or invert $\vA$ explicitly. Instead, we keep track of its upper triangular Cholesky factor $\vA = \vR\T \vR$~\cite{gijsberts2013real}. Given a new sample, a rank-1 update is applied to the Cholesky factor $\vR$, which requires $O(r^2)$ time. 
%
%
To cope with time-varying systems and to make the method more adaptive, we employ a forgetting factor $\lambda \in (0, 1)$, such that the impact of the previous samples decays exponentially in time~\cite{ljung1998system}. 
With this weighting criterion, we update $\vA$ and $\vPhi \Delta \vX$ with a new sample $(\tx, \Delta \vx)$ as \vspace{-2mm}
\begin{equation} \label{eqn_forget_factor_update}
\vA \leftarrow \lambda \vA + (1-\lambda)\vphi(\tx) \vphi(\tx)\T ,~~~~
\vPhi \Delta \vX \leftarrow \lambda \vPhi \Delta \vX+ (1-\lambda)\vphi(\tx) \Delta \vx\nonumber
\end{equation}
Here $\lambda \in (0, 1)$ is the forgetting factor~\cite{ljung1998system}, and $\vA$ and $\vPhi \Delta \vX$ are normalized by the number of offline training points $M$ after offline batch training $\vA \leftarrow \frac{1}{M} \vA$, $\vPhi \Delta \vX \leftarrow \frac{1}{M}\vPhi \Delta \vX$. This convex combination blends the previous samples and the current one. 


\vspace{-0.1 cm}
\subsection{Approximate Bayesian inference}\label{approximate inference}
\vspace{-0.1 cm}
When performing long-term prediction using the SSGP models,  the input state-control pair $\tx$ becomes uncertain. Here we define the joint distribution over state-control pair at one time step as $\rp(\tx)=\rp(\vx,\vu)$. Thus the distribution over state transition becomes 
$
\rp(\Delta\vx)  = \int\int\rp(\vf(\tx)|\tx)\rp(\tx)\rd\vf\rd\tx.
$
This predictive distribution cannot be computed analytically because the nonlinear mapping of an input Gaussian distribution leads to a non-Gaussian predictive distribution. Therefore we need to resort to approximate methods. Deterministic approximate inference methods 
approximate the posterior with a Gaussian~\cite{candela2003propagation,girard2003gaussian,deisenroth2014gaussian}, e.g., $\Delta\vx=\vf(\tx)\sim \mathcal {N}(\vmu_{\vf},\vSigma_{\vf})$.  However, these methods scale quadratically with the number of training samples, making them unsuitable for online learning with moderate amounts of data (e.g., 500 or more data points).

 In this section we present two approximate inference methods based on SSGPs. Our methods offer similar prediction performance compared with related methods in GPs. However, our methods scale quadratically with the number of \emph{random features}, which is usually much less than the number of training samples.
In this paper, we assume the conditional independency between different dimensions of $\Delta \vx$, given $\tx$, so in the following derivation we use $f$ to denote one of the dimensions. This assumption can be relaxed by using vector-valued kernels.

\subsubsection{Exact moment matching (SSGP-EMM)} \label{approximate_inf_EMM}
Given an input joint distribution $\mathcal{N}(\tmu,\tSigma )$, we may compute the exact posterior mean and variance. Applying the law of iterated expectation, the predictive mean $\mu_{f}$ is evaluated as
\footnotesize
\begin{equation*}
\mu_f = \E[f(\tx)|\tmu,\tSigma] = \E_{\tx}[\E_f[f(\tx)]] =\E_{\tx}[\vw\T \vphi(\tx)] = \vw \T \vq,
\quad \vq_i = \frac{\sigma_f}{\sqrt{r}} \;
\begin{cases}
\ec(\vomega_{i}) \quad
& i \leq r\\
\es(\vomega_{\bar i}) & i > r
\end{cases}
\end{equation*}
\normalsize
where 
\footnotesize
$\es(\tx) = \exp(-\frac{\tx\T \tSigma \tx}{2} )\sin(\tx\T \tmu)$,
$\ec(\tx) = \exp(-\frac{\tx\T \tSigma \tx}{2} )\cos(\tx\T \tmu)$,
\normalsize
and 
\footnotesize
$\bar i = \text{mod} (i, r)$ 
\normalsize
are defined for notation simplicity.
Next we compute the predictive variance  using the law of total variance
\scriptsize
\begin{equation}
\begin{split}
\Sigma_f &= \Var[f(\tx)|\tmu,\tSigma]\\
& = \E_{\tx}[\Var_f[f(\tx)]] + \Var_{\tx}[\E_f[f(\tx)]] \\
&= \E_{\tx}[\Var_f[f(\tx)]] + \E_{\tx}[\E_f[f(\tx)]^2] - \E_{\tx}[\E_f[f(\tx)]]^2 \\
&=  \E_{\tx}[\sigma_n^2 (1+ \vphi(\tx)\T\vA^{-1}\vphi(\tx))] + \E_{\tx}[(\vw\T\vphi(\tx))^2 ] - \mu_f^2 \\
&= \sigma_n^2 + \Tr \Big( 
\underbrace {\big(\sigma_n^2 \vA^{-1} + \vw \vw\T \big)}_\vS
\underbrace {\int \vphi(\tx) \vphi(\tx)\T p(\tx) \rd \tx}_\vT
\Big) - \mu_f^2 \nonumber
\end{split}~~~~~~
\begin{split}
\vT_{ij} =&
\frac{\sigma_f^2}{2r}  \cdot \;
\begin{cases}
+\ec(\vtheta_1) + \ec(\vtheta_2) \quad&
i\leq r , j \leq r
\\
+\es(\vtheta_1) - \es(\vtheta_2)&
i\leq r, j > r
\\
+\es(\vtheta_1) + \es(\vtheta_2)&
i > r, j \leq r
\\
-\ec(\vtheta_1) + \ec(\vtheta_2)&
i > r, j > r
\end{cases}
\\
\vtheta_1  =& \vomega_{\bar i} + \vomega_{\bar j} \quad \vtheta_2 = \vomega_{\bar i} - \vomega_{\bar j}
\end{split}
\end{equation}
\normalsize
The covariance between input and prediction can be computed as:
\scriptsize
\begin{equation}
\begin{split}
\vSigma_{\tx,f} &= \Cov [\tx, f(\tx) | \tmu, \tSigma] 
\approx \Cov_{\tx}[\tx, \E_f[f(\tx)]] 
=  \Cov_{\tx}[\tx, \vw\T \vphi(\tx)]  \\
&=  \E_{\tx}[(\tx-\tmu) (\vw\T \vphi(\tx) - \vw\T \vq)] \\
&=  \E_{\tx}[\vw \T \vphi(\tx) \tx - \vw \T \vq \tx - \vw\T \vphi(\tx) \tmu + \vw\T \vq \tmu ] \\ 
&=  \Big(\sum_{i=1}^{2r} \vw_i \underbrace{\E_{\tx}[\vphi_i(\tx) \tx]}_{\vP_i} \Big)
- \vw\T \vq \tmu 
- \vw\T \E_{\tx}[\vphi(\tx)]  \tmu + \vw\T \vq \tmu \\ 
&=  \vP  \vw  - \vw \T \vq \tmu
\end{split}~~~~~
\begin{split}
\vP_i = \frac{\sigma_f}{\sqrt{r}}  \cdot
\begin{cases}
\ec(\vomega_i)\tmu - \es(\vomega_i) \tSigma\vomega_i \quad
& i \leq r \\
\es(\vomega_{\bar i}) \tmu + \ec(\vomega_{\bar i}) \tSigma\vomega_{\bar i}
 & i > r
\end{cases}\nonumber
\end{split}
\end{equation}
\normalsize
Next we compute the covariance of two output dimensions (off-diagonal entries in the predictive covariance matrix) $f^i$, $f^j$ with uncertain input as follows:
\scriptsize
\begin{equation}
\begin{split}
\Cov[f^i, f^j |\tmu, \tSigma] &= \E_\tx [\E[f^i] \E[f^j]] -  
\E_\tx [\E[f^i]] \E_\tx [\E[f^j]] \\
&= \E_\tx [ ({\vw^i}\T \vphi^i) ( {\vw^j}\T \vphi^j)  ] -  \mu_f^i \mu_f^j \\
&= {\vw^i}\T  \E_\tx [ \vphi^i  {\vphi^j}\T  ]  {\vw^j}-  \mu_f^i \mu_f^j \\
&= {\vw^i}\T 
\underbrace{
\Big( \int \vphi^i {\vphi^j}\T p(\tx) \rd \tx\ \Big) }
_{\vT^{ij}}
{\vw^j} -  \mu_f^i \mu_f^j \\
\end{split}~~~~~~~~
\begin{split}
\vT^{ij}_{st} &=
\frac{\sigma_f^2}{2r}  \cdot \;
\begin{cases}
+\ec(\vtheta_1) + \ec(\vtheta_2) \quad&
s\leq r , t \leq r
\\
+\es(\vtheta_1) - \es(\vtheta_2)&
s \leq r, t > r
\\
+\es(\vtheta_1) + \es(\vtheta_2)&
s > r, t \leq r
\\
-\ec(\vtheta_1) + \ec(\vtheta_2)&
s > r, t > r
\end{cases}
\\
\vtheta_1  &= \vomega^i_{\bar s} + \vomega^j_{\bar t} \quad 
\vtheta_2 = \vomega^i_{\bar s} - \vomega^j_{\bar t}
\end{split}
\end{equation}
\normalsize
where superscript $i$ denotes the corresponding value related to the $i$th output, e.g. $\vw^i$, $\vphi^i$, and $\mu_f^i$ are the coefficients, features mapping, and predictive mean of the $i$th output, respectively. See the supplementary material for a detailed derivation of SSGP-EMM.

\subsubsection{Linearization (SSGP-Lin)}\label{approximate_inf_lin}
Another approach to approximate the predictive distribution under uncertain input is through linearizing of the posterior SSGP mean function. First we derive the partial derivative of the predictive mean $\E_f[f(\tx)]$ to the input $\tx$, and predictive mean's first-order Taylor expansion around the input mean
\begin{equation}
\frac{\partial \E_f[f(\tx)]}{\partial \tx} = 
\frac{\partial \big( \vw \T \vphi(\tx) \big)} {\partial \tx} =
\frac{\partial\vphi(\tx)} {\partial \tx} \vw =
\vD(\tx) \vw
\end{equation}
\normalsize
where
\footnotesize
$
\vD_i(\tx) = 
\frac{\sigma_f}{\sqrt{r}} \cdot
\begin{cases}
-\vomega_i \sin(\vomega_i\T\tx) \;
& i \leq r\\
+\vomega_{\bar i} \cos(\vomega_{\bar i}\T\tx) \;
& i > r
\end{cases}
$, \normalsize
is the $i$th column of $\vD(\tx)$.
\scriptsize
\begin{equation*}
\begin{split}
\hspace{-2mm}\E_f[f(\tx)] \approx \E_f[f(\tmu)] + \frac{\partial \E_f[f(\tx)]}{\partial \tx}\bigg|^{\rT}_{\tx=\tmu}\Big(\tx - \tmu\Big) =\vw\T \vphi(\tmu) + (\vD(\tmu)\vw)\ T(\tx- \tmu) = \underbrace{ (\vD(\tmu)\vw)\T}_{\va(\tmu)\T} \tx + \underbrace{\vw\T \vphi(\tmu) - (\vD(\tmu)\vw)\T \tmu}_{b(\tmu)}
\end{split}
\end{equation*}
\normalsize
The predictive mean $\mu_f$ is obtained  by evaluating the function at input mean $\tmu$. More precisely
\begin{equation}
\begin{split}
\mu_f = \E[f(\tx)|\tmu,\tSigma] =  \E_{f}[ \E_{\tx}[f(\tx)] ] 
\approx \E_f[f(\tmu)|\tmu] = \vw\T\vphi(\tmu)
\end{split}
\end{equation}
\normalsize
Based on the linearized model, the predictive variance $\Sigma_f$ with uncertain input is evaluated using the law of total variance
\begin{equation}
\begin{split}
\Sigma_f &= \Var [f(\tx)|\tmu,\tSigma]  = \Var_\tx [\E_f [f(\tx)]  ] + \E_\tx [\Var_f [f(\tx)] ]  \approx \Var_\tx [\va\T \tx + b ] + \Var_f[f(\tmu)] \\
&= \Var_\tx [\va\T \tx] + \Var_f[f(\tmu)] 
= \va\T \tSigma \;\va + \sigma_n^2\Big(1+ \vphi(\tmu)\T\vA^{-1}\vphi(\tmu)\Big) 
\end{split}
\end{equation}
\normalsize
The covariance between input and prediction $\vSigma_{\tx, f}$, and the covariance between two prediction can be computed as\vspace{-4mm}
\small
\begin{equation}
\begin{split}
\vSigma_{\tx,f} &= \Cov[\tx, f(\tx) | \tmu, \tSigma] \approx \Cov_{\tx}[\tx, \E_f[f(\tx)]]\nonumber\\ &=  \Cov_{\tx}[\tx, \va\T \tx + b]   \E_{\tx}[(\tx - \tmu) (\va\T \tx - \va\T \tmu)] =  \tSigma \va
\end{split}
\end{equation}
\begin{equation}
\begin{split}
\Cov[f^i, f^j |\tmu, \tSigma] 
&= \E_\tx \Big[ \Big(\E_f[f^i] - \E_\tx[\E_f[f^i]]\Big)  \Big(\E_f[f^j] - \E_\tx[\E_f[f^j]]\Big)\Big]\nonumber\\ &= \E_\tx [{\va^i}\T (\tx-\tmu) \;{\va^j}\T(\tx-\tmu)] ={\va^i}\T \tSigma \va^j
\end{split}
\end{equation}
\normalsize
Different from the exact moment matching used for SSGP inference, the approach presented here is an approximation of the posterior moments.  See Table~\ref{computational_complexity} for a comparison between our methods and exact moment matching approach for GPs (GP-EMM) \cite{candela2003propagation,girard2003gaussian,deisenroth2014gaussian}.

\begin{wraptable}{r}{7.cm} \vspace{-4mm}\scriptsize
    \begin{tabular}{| l | l | l | l |}
    \hline
     & Computational complexity \\ \hline
    GP-EMM \cite{candela2003propagation,girard2003gaussian,deisenroth2014gaussian} & $O\big(N^2 n(n+m)^2\big)$ \\ \hline
    SSGP-EMM & $O\big(r^2 n(n+m)^2\big)$   \\ \hline
    SSGP-Lin & $O\big(r^2 n + rn(n+m)  +n(n+m)^2\big)$  \\
    \hline
    \end{tabular}
    \caption{\footnotesize{Comparison of the proposed methods with the exact moment matching for GP. $n$ and $m$ are dimensions of state and control. $N$ is the number of training points. $r$ is the number of random features.}}\vspace{-4mm}\label{computational_complexity}
\end{wraptable}

\paragraph{Belief space representation} Given the above expressions,  we can compute the predictive distribution $\vmu_{k+1}, \vSigma_{k+1}$  as follows
\begin{equation}\label{gpdyn}
\begin{split}
\vmu_{k+1} &= \vmu_k + \vmu_{\vf_k} \\
 \vSigma_{k+1} &= \vSigma_k + \vSigma_{\vf_k} + \vSigma_{\tx_k,\vf_k} + \vSigma_{\vf_k,\tx_k}.
\end{split}
\end{equation} 
\normalsize
Note that $\vmu_{\vf_k},\vSigma_k$ and $\vSigma_{\tx_k,\vf_k}$ are nonlinear functions of  $\vmu_k $ and $\vSigma_k $.  We define the belief as the predictive distribution $\vb_k=[\vmu_k~ \text{vec}(\vSigma_k)]^{\rT}$ over state $\vx_k$, where $\text{vec}(\vSigma_k) $ is the vectorization of $\vSigma_k$. Therefore eq (\ref{gpdyn}) can be written in a compact form
\begin{equation}
\vb_{k+1} = \mF(\vb_k,\vu_k),
\end{equation}
where $\mF$ is defined by (\ref{gpdyn}). The above equation corresponds to the belief space representation of the unknown dynamics in eq (\ref{real_dyn}) in discrete-time.

\vspace{-0.1 cm}
\section{Online Probabilistic Trajectory Optimization}
\vspace{-0.1 cm}
In order to incorporate dynamics model uncertainty explicitly, we perform trajectory optimization in belief space.  In our proposed framework,  at each iteration we create  a local model along a nominal trajectory through the belief space $(\bar{\vb}_k,\bar{\vu}_k) $ including i)  a linear approximation of the belief dynamics model; ii) a second-order local approximation of the  value function. We define the belief and control nominal trajectory ($\bar{\vb}_{1:H},\bar{\vu}_{1:H}$) and deviations from this trajectory $\delta\vb_k=\vb_k-\bar{\vb}_k$, $\delta\vu_k=\vu_k-\bar{\vu}_k$.  The  linear approximation of the belief dynamics along the nominal trajectory is
\footnotesize
\begin{align}\label{belief_dyn}
\delta\vb_{k+1} = \left[ \begin{array}{cc}
\frac{\partial \vmu_{k+1}}{\partial \vmu_k} & \frac{\partial \vmu_{k+1}}{\partial \vSigma_k} \\
\frac{\partial \vSigma_{k+1}}{\partial \vmu_k}     &   \frac{\vSigma_{k+1}}{\partial \vSigma_k} \end{array} \right]\delta\vb_k +  \left[ \begin{array}{c}  \frac{\partial \vmu_{k+1}}{\partial \vu_k} \\
\frac{\partial \vSigma_{k+1}}{\partial \vu_k}\end{array} \right]\delta\vu_k,
\end{align}
\normalsize  
All partial derivatives  are computed analytically. 
Based on the dynamic programming principle, the value function is the solution to the Bellman equation
\footnotesize
 \begin{align}\label{eq_bellman}
V(\vb_k,k) = \min_{\vu_k}\big(\underbrace{\mathcal{L}(\vb_k,\vu_k) + V\big(\vf(\vb_k,\vu_k),k+1 \big)}_{Q(\vb_k,\vu_k)}\big).
\end{align}
\normalsize
The $Q$ function can be approximated as a quadratic model along the nominal trajectory \cite{jacobson1970differential}. The local optimal control law is computed by minimizing the approximated $Q$ function
\footnotesize
\begin{equation}\label{policy}
\delta\hat{\vu}_k = \arg\min_{\delta\vu_k} \big[Q_k(\vb_k+\delta \vb_k,\vu_k+\delta \vu_k) \big] = -(Q^{uu}_k)^{-1}Q_k^u -(Q^{uu}_k)^{-1}Q_k^{ux}\delta\vb_k . 
\end{equation}
\normalsize
where superscripts of $Q$ indicate partial derivatives.   The new control is obtained as $\hat{\vu}_k=\bar{\vu}_k+\delta\hat{\vu}_k$, and the quadratic approximation of the value function is propagated backward in time iteratively. See the supplementary material for details regarding the backward propagation.  

We use the learned control policy to generate a locally optimal trajectory by propagating the belief dynamics forward in time  using the proposed approximate inference methods, i.e., SSGP-EMM (\ref{approximate_inf_EMM}) or SSGP-Lin (\ref{approximate_inf_lin}). Note that the belief dynamcis model is determinsitc. And the instantaneous cost $\mathcal{L}(\vv_k,\vu_k)=\mathbb{E}[\mathcal{L}(\vx_k,\vu_k)| \vmu_k,\vSigma_k]$. In our implementation we apply line search to   guarantee convergence to a locally optimal control policy \cite{todorov2005generalized}. Control constraints are taken into account using the method in \cite{tassacontrol}. The summary of algorithm is included in the supplementary material.
\vspace{-0.1 cm}
\subsection{Online optimization: a receding horizon scheme}
\vspace{-0.1 cm}
The trajectory optimization approach with learned dynamics can be used for episodic RL. However, this requires interactions with the physical system over a long time scale (trajectory), and is not reactive to task or model variations that occur in short time scales (e.g., at every time step).   Here we propose an online approach in the spirit of  model predictive control (MPC).

Given a  solution to a single H-step trajectory optimization problem starting at $\vx_1$, we only apply the first element of the control sequence $\vu_1$ and proceed to solve another H-step trajectory optimization problem starting at $\vx_2$. Different from the offline approach, we initialize with the previous optimized trajectory. The H-step nominal control sequence for warm-start becomes $\vu_2,\vu_3,...,\vu_H,\vu_H$. The online optimization would converge much faster than the offline case as long as the new target is not far from the previous one because of the warm-start. In addition, at $\vx_2$ we collect the training pair $\{(\vx_1,\vu_1),\Delta\vx_1\}$ and update the learned SSGP model as introduced in section \ref{SSGP}. 

\vspace{-0.3 cm}
\begin{algorithm}[H]
\footnotesize
\caption{\footnotesize{Adaptive Probabilistic Trajectory Optimization \label{algorithm_oline} (1-3: \textbf{offline learning}, 4-8: \textbf{online learning}})}
\begin{algorithmic}[1]
\State \textbf{Initialization:}  Collect offline data.
\State \textbf{Model learning:} Train GP hyperparameters. Sample random features and compute their weights (sec.\ref{SSGP}).
\State \textbf{Trajectory optimization:} Perform belief space DDP based on the learned model (algorithm 1 in supplementary material).
\Repeat
\State \parbox[t]{\dimexpr\linewidth-\algorithmicindent}{ \textbf{Policy execution:}  Apply one-step control $\hat{\vu}_1$ to the system and move one step forward. Record data. \strut}
\State \parbox[t]{\dimexpr\linewidth-\algorithmicindent}{  \textbf{Model adaptation:} Incorporate data and update random features' weights $\vw$ (sec.\ref{SSGP}). \strut}
\State \parbox[t]{\dimexpr\linewidth-\algorithmicindent}{  \textbf{Trajectory optimization:} Perform re-optimization with  the updated model (algorithm 1 in supplementary material). Initialize with the previously optimized policy/trajectory. \strut}
\Until Task terminated 
\end{algorithmic}
\normalsize
\end{algorithm}

\vspace{-0.5 cm}

This online scheme is particularly suitable for applications with task or model variations. In contrast, most state of the art RL methods could not be efficiently applied in those cases. An summary of the algorithm can be found in algorithm \ref{algorithm_oline}.

\vspace{-0.1 cm}
\subsection{Relation to prior work}
\vspace{-0.1 cm}
Our method shares some similarities with methods such as  iLQG-LD\cite{mitrovic2010adaptive}, AGP-iLQR \cite{boedecker2014approximate},  PDDP\cite{pan2014probabilistic} and Minimax DDP\cite{morimoto2002minimax} . All of these methods are based on DDP or iLQR and learned dynamics models. Our method differs 
in both model and controller learning. 
GP models  are more robust to modeling error than  LWPR (iLQG-LD \cite{mitrovic2010adaptive}) or RFWR (Minimax DDP \cite{morimoto2002minimax}). But the major obstacle for applying GPs is the high computational demand for performing inference. Our method integrates scalable approximate inference (see section \ref{approximate inference}) into trajectory optimization. In contrast, related methods either employs computation-intensive inference approach (PDDP \cite{pan2014probabilistic}), or drops uncertainty in the dynamics model (AGP-iLQR \cite{boedecker2014approximate}), which becomes less robust to modeling errors.  
Our method also performs fast online model adaptation and re-optimization. These features are essential when we are dealing with 1) condition variations, such as time-varying tasks, dynamics or environment; 2) infinite horizon problems, such as stabilization.


\vspace{-.1 cm}
\section{Experiments}
\vspace{-0.1 cm}
\vspace{-0.1 cm}
\subsection{Approximate inference performance}\label{sec_exp_inf}
\vspace{-0.1 cm}
We compare the proposed approximate inference methods with three existing approaches: the full GP exact moment matching (GP-EMM) approach \cite{candela2003propagation,girard2003gaussian,deisenroth2014gaussian}, Subset of Regressors GP (SoR-GP)  \cite{williams2006gaussian} used in AGP-iLQR \cite{boedecker2014approximate},   and LWPR \cite{vijayakumar2005incremental} used in iLQG-LD \cite{mitrovic2010adaptive}. Note that SoR-GP and LWPR do not take into account input uncertainty when performing regressions. We consider two multi-step prediction tasks using the dynamics models of a quadrotor  (16 state dimensions, 4 control dimensions) and a Puma-560 manipulator (12 state dimensions, 6 control dimensions).

\begin{figure*}[!htb]
         \begin{subfigure}[b]{0.25\textwidth}
             \includegraphics[width=\textwidth]{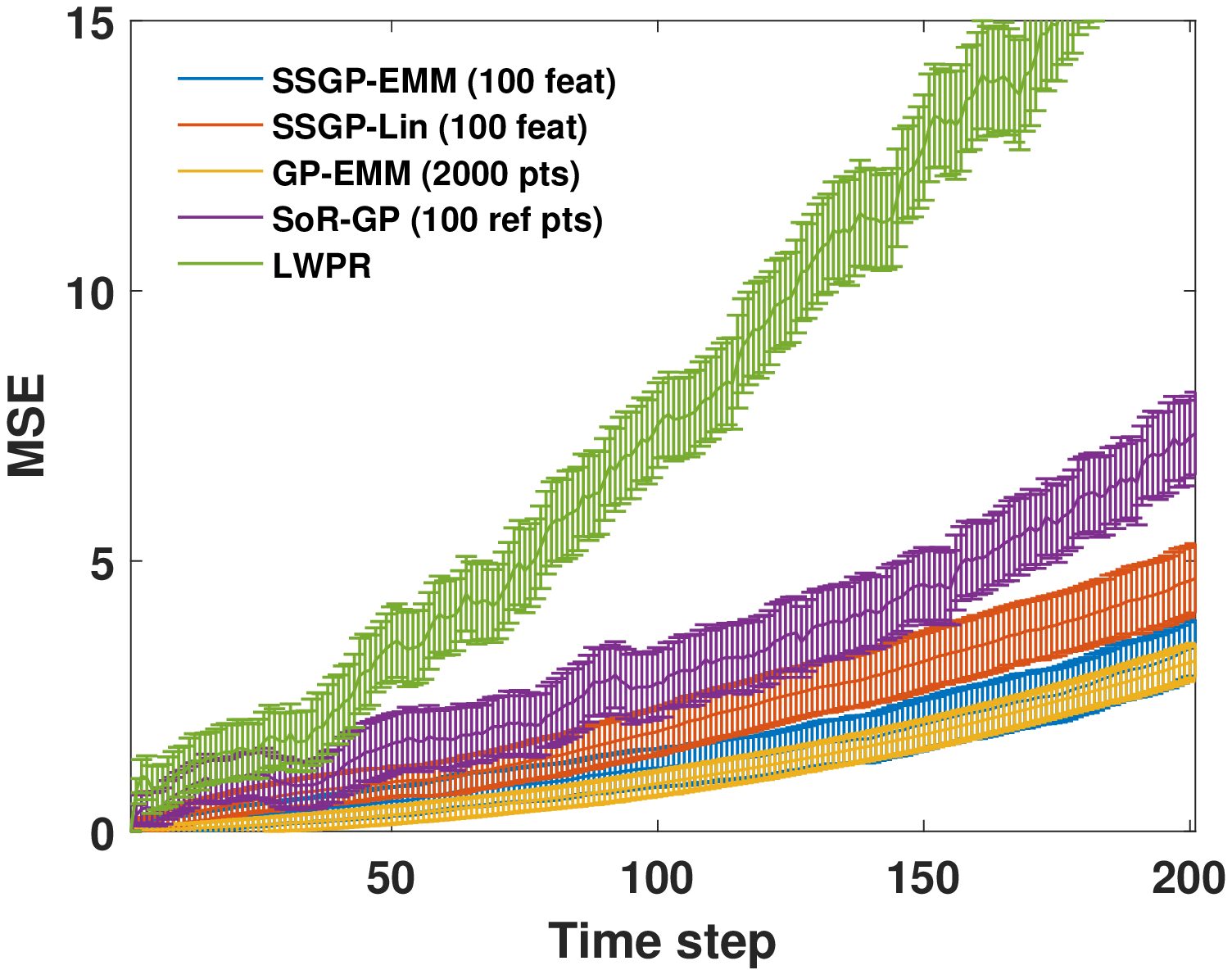}
              \caption{Quadrotor}
              \label{fig_time_lin}
        \end{subfigure}%
                 \begin{subfigure}[b]{0.25\textwidth}
             \includegraphics[width=\textwidth]{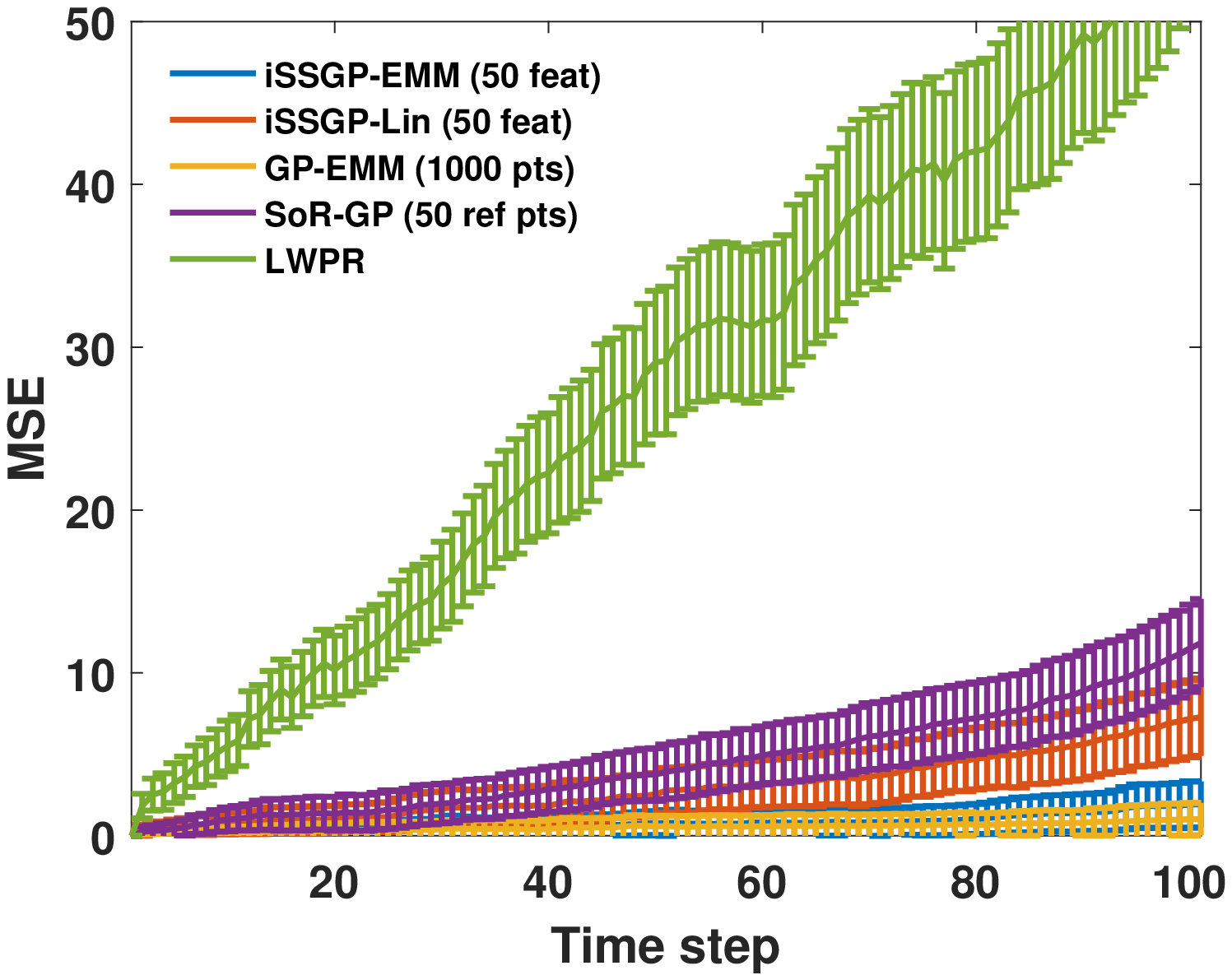}
              \caption{Puma 560}
              \label{fig_time_lin}
        \end{subfigure}%
         \begin{subfigure}[b]{0.25\textwidth}
                \includegraphics[width=\textwidth]{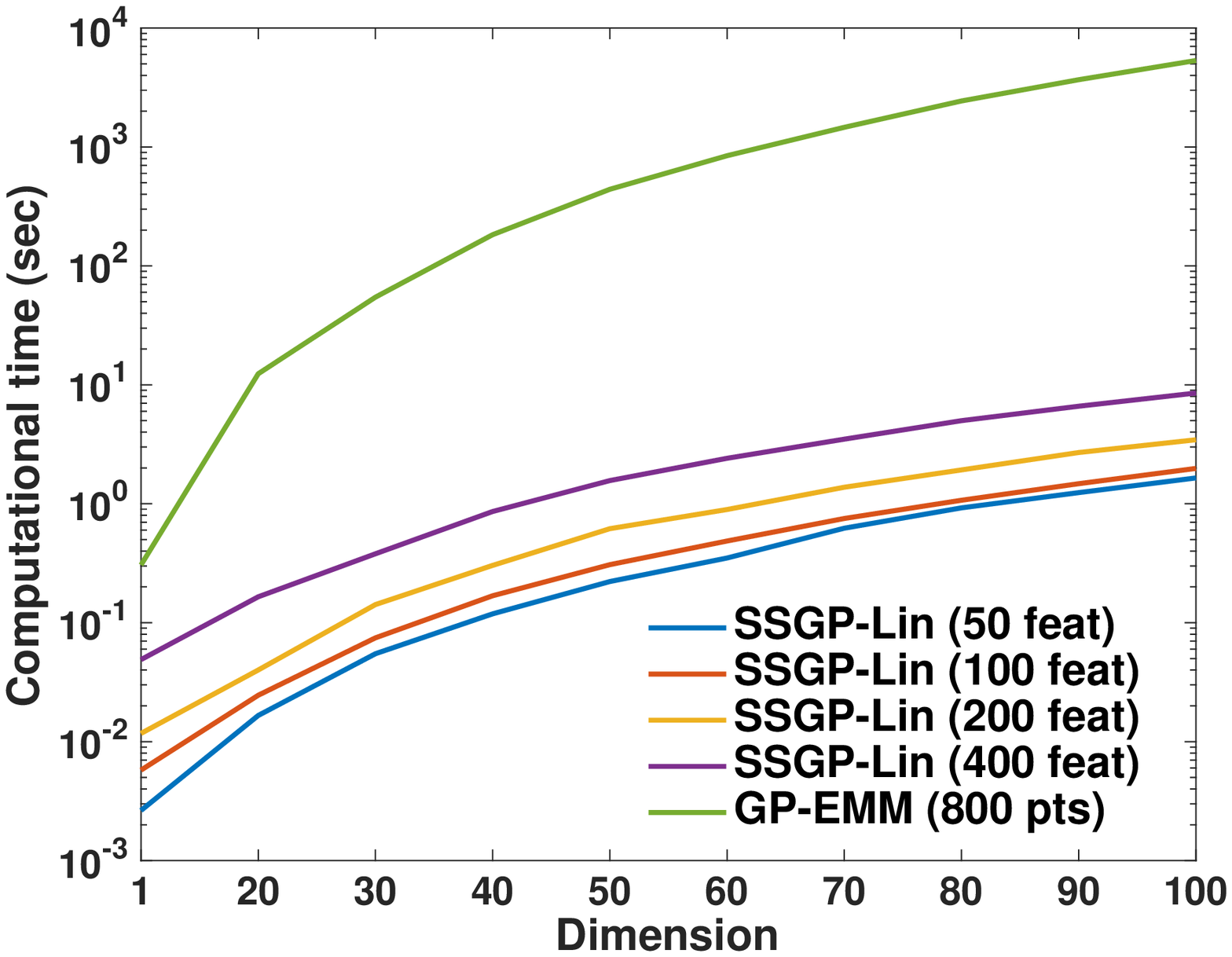}
                \caption{Computation time}
                \label{fig_time_lin}
        \end{subfigure}%
        \begin{subfigure}[b]{0.25\textwidth}
                \includegraphics[width=\textwidth]{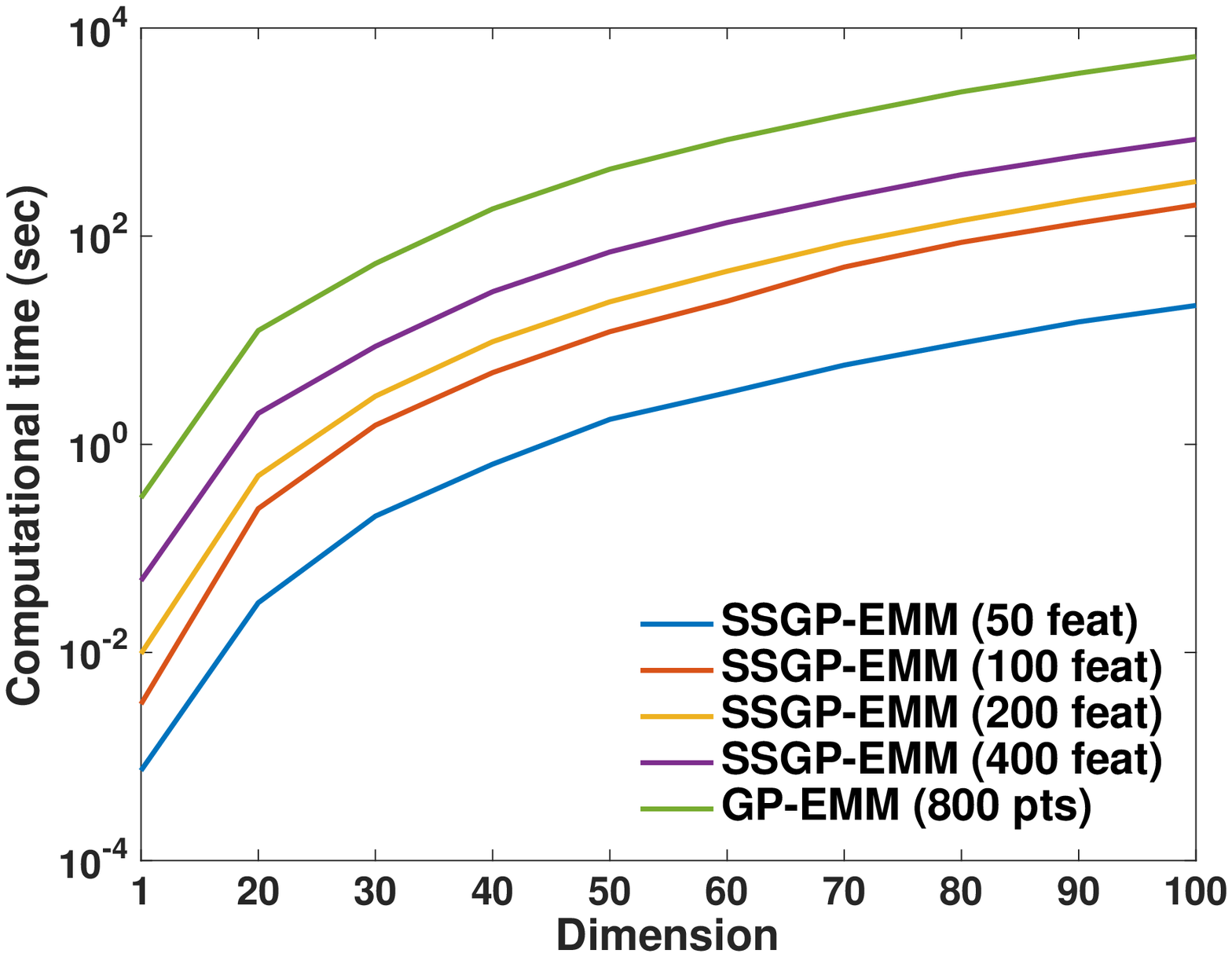}
                \caption{Computation time}
                \label{fig_time_emm}
        \end{subfigure}
        \caption{(a)-(b): Approximate inference accuracy test. The vertical axis is the squared error of cost predictions for (a) quadrotor system and (b) Puma 560 system. Error bars represent standard deviations over 10 independent rollouts. (c)-(d): Comparison of computation time on a log scale  between (c) SSGP-Lin and GP-EMM; (d) SSGP-EMM and GP-EMM. The horizontal axis is the input and output dimension (equal in this case). Vertical axis is the CPU time in seconds.}\vspace{-3mm}\label{approx_inf}
\end{figure*}

\paragraph{Accuracy of multi-step prediction} In the following, we evaluate the performance in terms of prediction accuracy.  We collected training sets of 1000 and 2000 data points for the quadrotor and puma task, respectively. For model learning we used 100/50 random features for our methods and 100/50 reference points for SoR-GP. Based on the learned models, we used a set of 10 initial states and control sequences to perform rollouts (200 steps for quadrotor and 100 steps for Puma) and compute the cost expectations at each step.  Fig.\ref{approx_inf}(a)(b) shows the cost prediction errors, i.e.$(\mathcal{L}(\tx_k)-\E[\mathcal{L}(\tx_k)|\tmu_k,\tSigma_k])^2$. It can be seen that SSGP-EMM is very close to GP-EMM and SSGP-EMM performs slightly better than SSGP-Lin in all cases. Since SoR-GP and LWPR do not take into account input uncertainty when performing regression, our methods outperform them consistently.

\paragraph{Computational efficiency}
In terms of the computational demand, we tested the CPU time for one-step prediction using SSGP-EMM and SSGP-Lin and full GP-EMM. We used sets of 800 random data points of 1,10,20,30,40,50,60,70,80,90 and 100 dimensions to learn SSGP and GP models. The results are shown in fig.\ref{fig_time_lin},\ref{fig_time_emm}. Both SSGP-EMM and SSGP-Lin show  significant less computational demand than GP-EMM. In contrast, as shown in the last subsection and fig.\ref{approx_inf}, their prediction performances differences are not substantial. See section 3 for a comparison in terms of computational complexity. Our methods are more scalable than GP-EMM, which is the major computational bottleneck for probabilistic model-based RL approaches \cite{deisenroth2014gaussian,pan2014probabilistic}.

\vspace{-0.1 cm}
\subsection{Trajectory optimization performance}
\begin{figure}[!htb]
        \begin{subfigure}[b]{0.3\textwidth}
                \includegraphics[width=\textwidth]{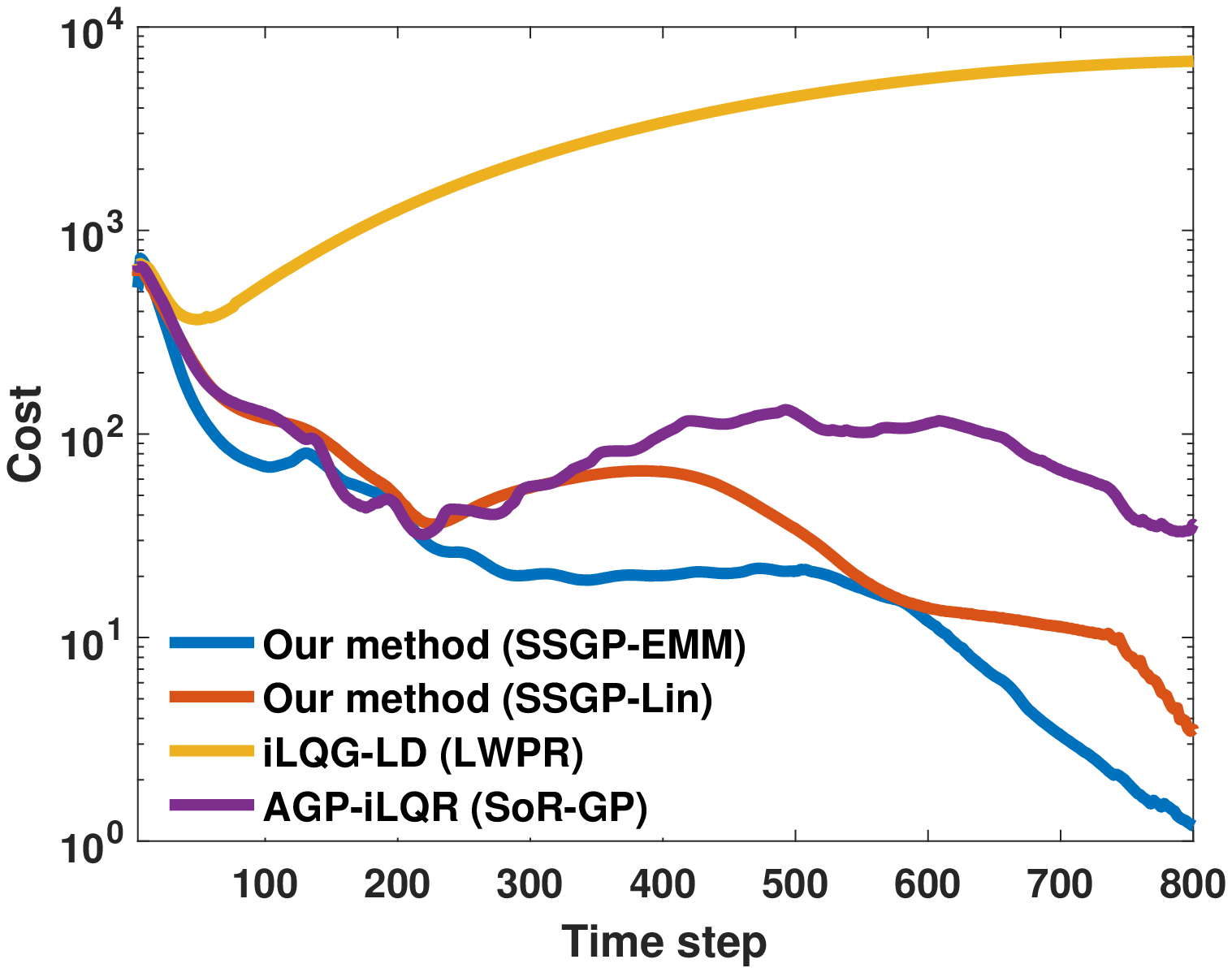}
                \caption{Puma 560 cost comparison }
                \label{fig_p560_cost}
        \end{subfigure}%
        \begin{subfigure}[b]{0.18\textwidth}
                \includegraphics[width=\textwidth]{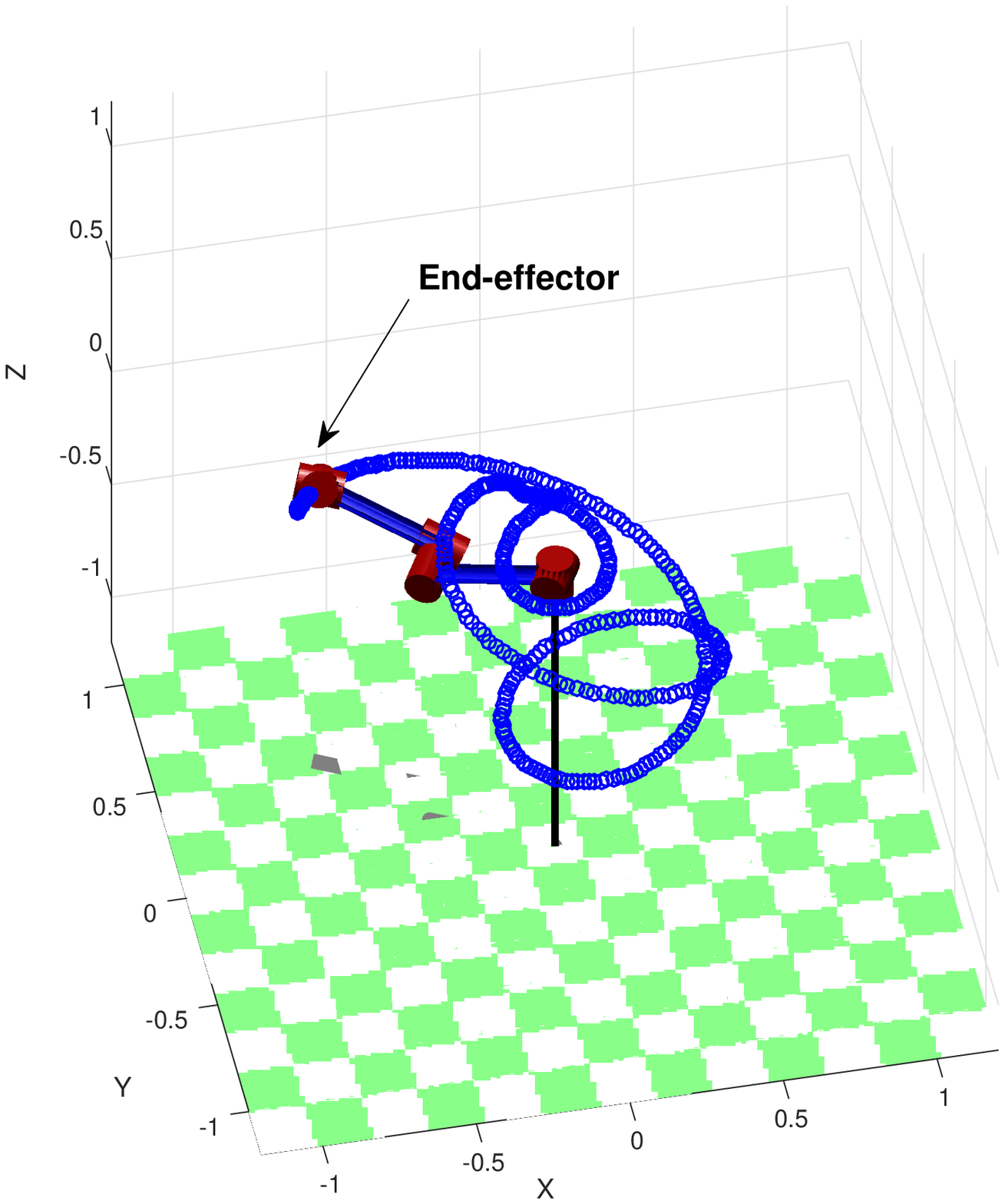}
                \caption{Puma 560 task}
                \label{fig_p560_target}
        \end{subfigure}
           \begin{subfigure}[b]{0.33\textwidth}
                \includegraphics[width=\textwidth]{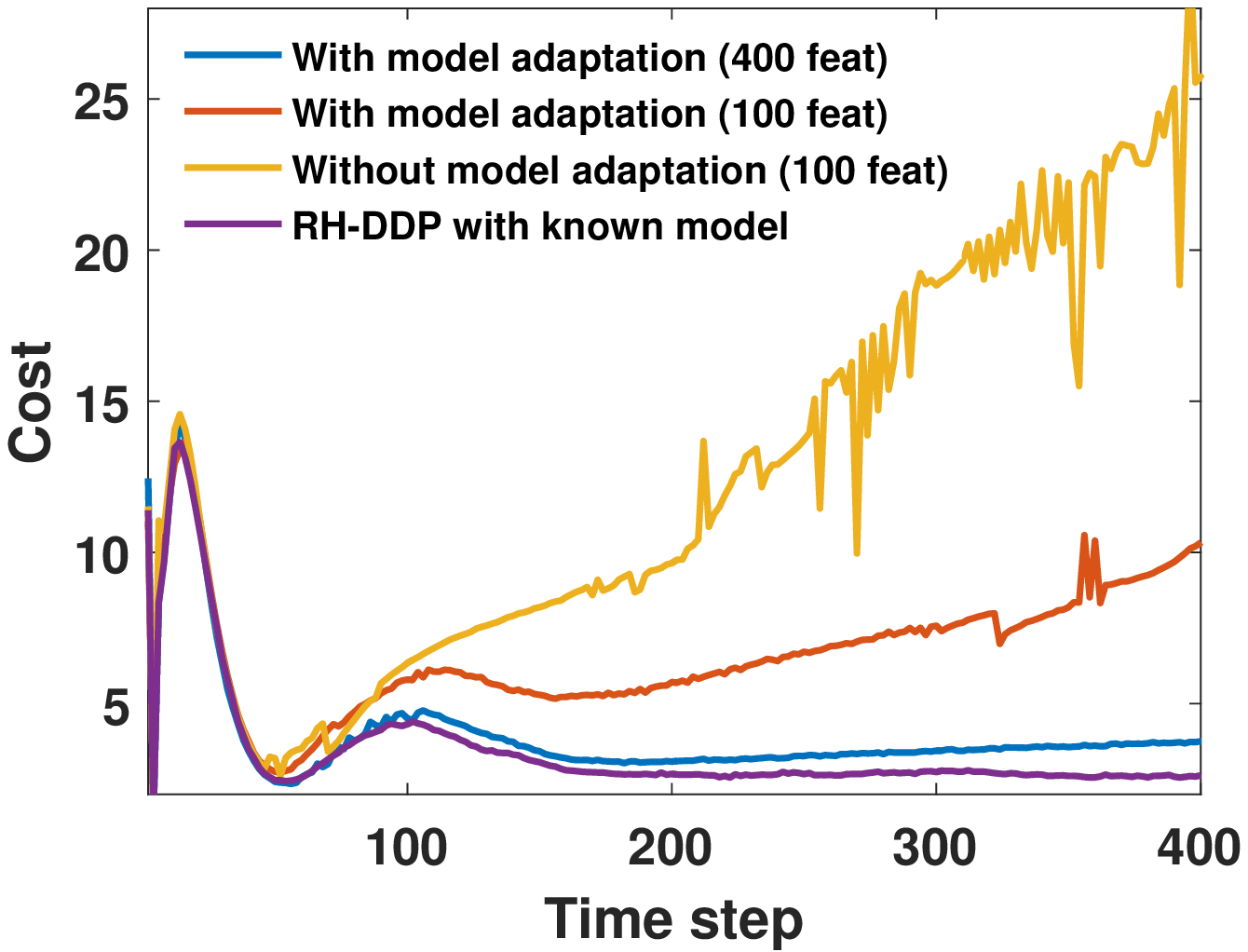}
                \caption{Quadrotor cost comparison}
                \label{fig_quad_cost}
        \end{subfigure}%
        \begin{subfigure}[b]{0.26\textwidth}
                \includegraphics[width=\textwidth]{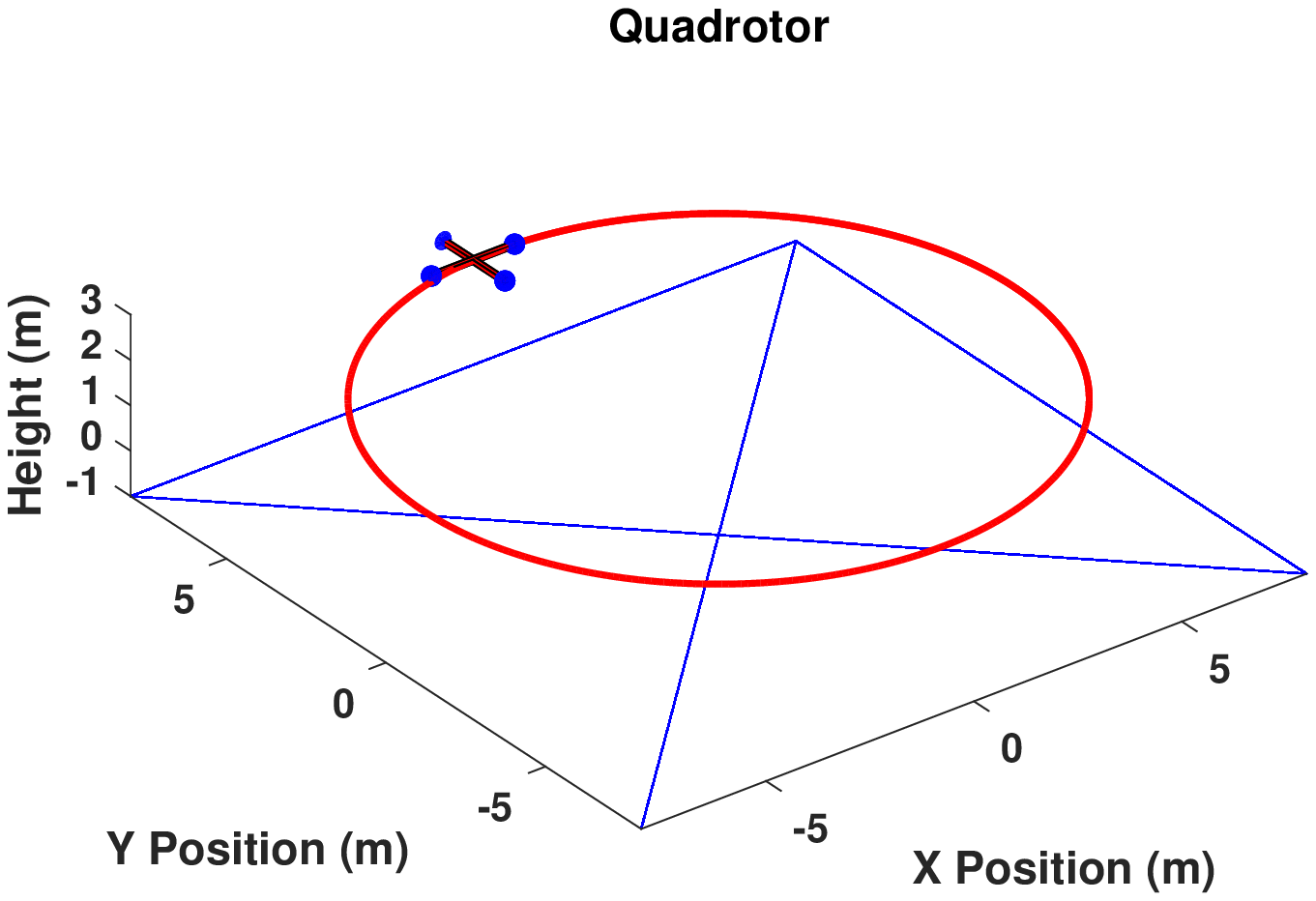}
                \caption{Quadrotor task}
                \label{fig_quad_traget}
        \end{subfigure}
        \caption{Trajectory optimization performances and tasks}\vspace{-4mm}\label{fig_p560}
\end{figure}

\vspace{-0.1 cm}
We evaluate the performance of our methods using three model predictive control (MPC) tasks.
\paragraph{PUMA-560 task: moving target and model parameter changes}
The task is to steer the end-effector to the desired position and orientation. The desired state is time-varying over 800 time steps as shown in fig.\ref{fig_p560_target}.
We collected 1000 data points offline and sampled 50 random features for both of our methods. Similarly for AGP-iLQR we used 50 reference points.  In order to show the effect of online adaptation, we increased the mass of the end-effector by 500$\%$ at the beginning of online learning (it is fixed during learning).  Fig.\ref{fig_p560}(a) shows the cost reduction results averaged over 3 independent trials.  Our method based on SSGP-EMM slightly outperforms SSGP-lin based method. Although iLQG-LD and AGP-iLQR also feature online model adaptation and DDP-based optimization, their controls are computed based on predictions that are more biased than our methods (see comparisons in section \ref{sec_exp_inf}). As a result our methods show superior predictive control performance.

\paragraph{Quadrotor task: time-varying tasks and dynamics}
The objective is to start at  (-1, 1, 0.5) and track a moving target as shown in fig.\ref{fig_quad_traget} for 400 steps. The mass of the quadrotor is decreasing at a rate of 0.02 kg/step. The controls are thrust forces of the 4 rotors and we consider the control constraint $\vu_{\min}=0.5, \vu_{\max}=3$. We collected 3000 data points offline, and sampled 100 and 400 features for online learning. The forgetting factor for online learning $\lambda=0.992$. SSGP-Lin was used for 
approximate inference.  Results are shown in fig. \ref{fig_quad_cost}. The effect of online model adaptation is significant after 100 steps due to the time-varying dynamics. Not surprisingly, increasing the number of features results in better performance. The receding-horizon DDP (RH-DDP) \cite{tassa2007receding} with full knowledge of the dynamics model was used as a baseline.

\begin{figure}[!htb]
\centering
\includegraphics[width=1.0\textwidth]{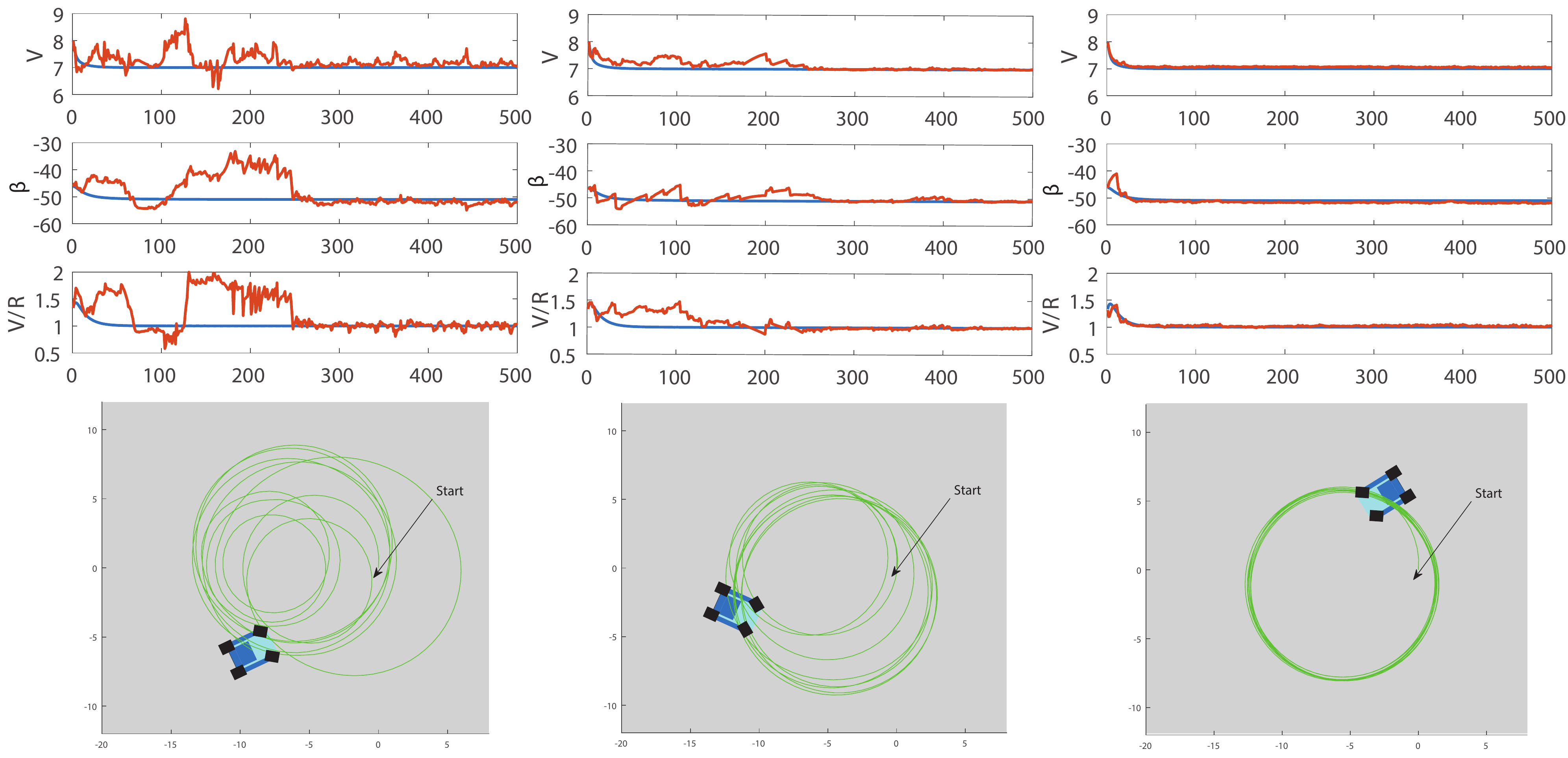}
\caption{Comparison of the stabilization performance using (1st column) 50, (2nd column) 150 and (3rd column) 400 random features. Blue lines are the analytic LQR solution in \cite{velenis2010steady}}\label{fig_rc}
\end{figure}

\paragraph{Autonomous driving during extreme conditions: steady-state stabilization}
In this example, we study the control of a wheeled vehicle during extreme operation conditions (powerslide). The task is to stabilize the vehicle to a specified steady-state  using purely longitudinal control. The desired steady-state consists of velocity $V$ , side slip angle $\beta$, and yaw rate $\frac{V}{R}$ where $R$ is the path radius. This problem has been studied in \cite{velenis2010steady} where the authors developed a LQR control scheme based on analytic linearization of the dynamics model. However, this method is restrictive due to the assumption of full knowledge of the complex dynamics model. We applied our method to this task under unknown dynamics with 2500 offline data points, which were sampled from the empirical vehicle model in \cite{velenis2010steady}. We used 50, 150, and 400 random features and SSGP-EMM for approximate inference in our experiments. Results and comparisons with the solution in  \cite{velenis2010steady} are shown in fig.\ref{fig_rc}.  As can be seen, in the case of 400 random features, our solution is very close to the analytic LQR solution and the system is stabilized after 30 time steps. With only 50 features, our method is sill able to stabilize the system after 300 time steps. Our method is suitable for infinite horizon control tasks due to its feature of efficient learning and online optimization.
\vspace{-0.2 cm}
        
\vspace{-0.1 cm}
\section{Conclusion} \label{sec:conclusion}
\vspace{-0.1 cm}
This paper presents a trajectory optimization framework for solving optimal control problems under uncertain dynamics. Similar to RL, our method can efficiently learn from experience and adapt to new situations. Different from most RL algorithms, our method updates control policy and dynamics model in an online incremental fashion under  tasks and dynamics variations.  In order to perform robust and fast planning, we have introduced two scalable approximate inference methods. Both methods have demonstrated  superior  performance in terms of computational efficiency and long-term prediction accuracy compared to well-known methods. Our  adaptive probabilistic trajectory optimization framework combines the benefits of efficient inference and model predictive control (MPC), and is applicable  to a wide range of learning and control problems. 
Future work will include  1) partially observable learning and  2) transfer learning using samples from other models.

\bibliography{references}

\newpage

\title{Supplementary Material for Adaptive Probabilistic Trajectory Optimization via Efficient Approximate Inference}
\author[Yunpeng Pan et al.]
       {\textbf{Yunpeng Pan$^{1,2}$, Xinyan Yan$^{1,3}$, Evangelos Theodorou$^{1,2}$, and Byron Boots$^{1,3}$}\\
       $^1$Institute for Robotics and Intelligent Machines, Georgia Institute of Technology\\
       $^2$School of Aerospace Engineering, Georgia Institute of Technology \\ $^3$School of Interactive Computing, Georgia Institute of Technology \\
       \{ypan37,xyan43,evangelos.theodorou\}@gatech.edu, bboots@cc.gatech.edu\\ 
       }

%

\newcommand{\fix}{\marginpar{FIX}}
\newcommand{\new}{\marginpar{NEW}}

\setlength{\textfloatsep}{1pt} 
\setlength{\abovecaptionskip}{0pt}
\setlength{\belowcaptionskip}{0pt}


\maketitle


\section{Introduction}
In this supplementary note we provide  detailed derivation for the exact moment matching  and linearization methods presented in the main paper. We also provide the main equations for Differential Dynamic Programming. 

\section{Detailed Derivation of  Moment Matching Method }
Below we provide identities that  are important for the derivation of Moment Matching method. In particular, we consider  $\vx \in \vR^n$ and we will have:
\begin{equation} \label{eqn_quad_gaussian_integral}
\int \exp(-\vx\T\vA\vx + \vv\T \vx) ~\rd \vx = \pi^{\frac{n}{2}} |\vA|^{-\frac{1}{2}} \exp(\frac{1}{4}\vv\T \vA^{-1} \vv) =: \eta
\end{equation}

\begin{equation} 
\int (\va\T \vx)  \exp(-\vx\T\vA\vx + \vv\T \vx) ~\rd \vx = \va\T (\frac{1}{2}\vA^{-1}\vv) \eta
\end{equation}

\begin{equation} \label{eqn_multivariant_gaussian}
p(\vx) = (2\pi)^{-\frac{n}{2}} |\vSigma|^{-\frac{1}{2}} \exp (-\frac{1}{2} (\vx - \vmu)\T \Sigma ^{-1} (\vx - \vmu))
 \quad \vx \sim \mathcal{N} (\vmu, \vSigma)
\end{equation}

Consequently, we can derive, $\vx \sim \mathcal{N}(\vmu,\vSigma)$:
\begin{equation} \label{eqn_cos_gaussian_integral}
\int \cos(\vomega\T \vx) ~p(\vx) ~\rd \vx = \exp\big(-\frac{\vomega\T\vSigma\vomega}{2}\big)\cos(\vomega\T\vmu)
\end{equation}

\begin{equation} \label{eqn_sin_gaussian_integral}
\int \sin(\vomega\T \vx) ~p(\vx) ~\rd \vx = \exp\big(-\frac{\vomega\T\vSigma\vomega}{2}\big)\sin(\vomega\T\vmu)
\end{equation}

\begin{equation} \label{eqn_x_cos_gaussian_integral}
\begin{split}
\int \cos(\vomega\T \vx) \vx ~p(\vx) ~\rd \vx 
&= \exp\big(-\frac{\vomega\T\vSigma\vomega}{2}\big)\cos(\vomega\T\vmu) \vmu \\
&- \exp\big(-\frac{\vomega\T\vSigma\vomega}{2}\big)\sin(\vomega\T\vmu) \vSigma\vomega
\end{split}
\end{equation}

\begin{equation} \label{eqn_x_sin_gaussian_integral}
\begin{split}
\int \cos(\vomega\T \vx) \vx ~p(\vx) ~\rd \vx 
&= \exp\big(-\frac{\vomega\T\vSigma\vomega}{2}\big)\sin(\vomega\T\vmu) \vmu \\
&+ \exp\big(-\frac{\vomega\T\vSigma\vomega}{2}\big)\cos(\vomega\T\vmu) \vSigma\vomega
\end{split}
\end{equation}

\begin{align} \label{eqn_cos_cos_gaussian_integral}
\int \cos(\vomega_i\T \vx) \cos(\vomega_j\T \vx) ~p(\vx) ~\rd \vx
=& \frac{1}{2}\exp\big(-\frac{\vtheta_1\T\vSigma\vtheta_1}{2}\big)\cos(\vtheta_1\T\vmu) \\ +
&\frac{1}{2} \exp\big(-\frac{\vtheta_2\T\vSigma\vtheta_2}{2}\big)\cos(\vtheta_2\T\vmu) \quad
\vtheta_1 := \vomega_i + \vomega_j, \vtheta_2 := \vomega_i - \vomega_j \nonumber
\end{align}

\begin{align} \label{eqn_sin_sin_gaussian_integral}
\int \sin(\vomega_i\T \vx) \sin(\vomega_j\T \vx) ~p(\vx) ~\rd \vx
=& \frac{1}{2}\exp\big(-\frac{\vtheta_1\T\vSigma\vtheta_1}{2}\big)\cos(\vtheta_1\T\vmu) \\ -
&\frac{1}{2} \exp\big(-\frac{\vtheta_2\T\vSigma\vtheta_2}{2}\big)\cos(\vtheta_2\T\vmu) \quad
\vtheta_1 := \vomega_i - \vomega_j, \vtheta_2 := \vomega_i + \vomega_j \nonumber
\end{align}

\begin{align} \label{eqn_cos_sin_gaussian_integral}
\int \cos(\vomega_i\T \vx) \sin(\vomega_j\T \vx) ~p(\vx) ~\rd \vx
=& \frac{1}{2}\exp\big(-\frac{\vtheta_1\T\vSigma\vtheta_1}{2}\big)\sin(\vtheta_1\T\vmu) \\ -
&\frac{1}{2} \exp\big(-\frac{\vtheta_2\T\vSigma\vtheta_2}{2}\big)\sin(\vtheta_2\T\vmu) \quad
\vtheta_1 := \vomega_i + \vomega_j, \vtheta_2 := \vomega_i - \vomega_j \nonumber
\end{align}

The proof of Eq.~\ref{eqn_cos_gaussian_integral}:
\begin{align*}
&\int \cos(\vomega\T \vx) ~p(\vx) ~\rd \vx, \quad \vx \sim \mathcal{N}(\vmu,\vSigma)\\
= &\Re \Big\{\int (\cos(\vomega\T \vx) + i\sin(\vomega\T \vx)) ~p(\vx) ~\rd \vx \Big\}\\
= &\Re \Big\{\int \exp(i \vomega\T \vx)  (2\pi)^{-\frac{n}{2}} |\vSigma|^{-\frac{1}{2}} 
\exp (-\frac{1}{2}(\vx - \vmu)\T \vSigma^{-1} (\vx - \vmu)) \rd \vx
\Big\} \quad \quad\text{(Eq.~\ref{eqn_multivariant_gaussian})} \\
= &\Re \Big\{\int \exp(i\vomega\T \vmu) \exp(i \vomega\T (\vx-\vmu))  (2\pi)^{-\frac{n}{2}} |\vSigma|^{-\frac{1}{2}} 
\exp (-(\vx - \vmu)\T (2\vSigma)^{-1} (\vx - \vmu)) \rd \vx
\Big\} \\
= & (2\pi)^{-\frac{n}{2}} |\vSigma|^{-\frac{1}{2}} 
\Re \Big\{
\exp(i\vomega\T \vmu)
\int \exp (-(\vx - \vmu)\T (2\vSigma)^{-1} (\vx - \vmu) + (i \vomega)\T (\vx-\vmu))
\rd \vx
\Big\} \\
= & (2\pi)^{-\frac{n}{2}} |\vSigma|^{-\frac{1}{2}} 
\Re \Big\{
\exp(i\vomega\T \vmu)
\pi^{\frac{n}{2}} |\frac{1}{2}\vSigma^{-1}|^{-\frac{1}{2}} \exp(\frac{1}{4} (i\vomega)\T 2 \vSigma (i \vomega))
\Big\} 
\quad \quad\text{(Eq.~\ref{eqn_quad_gaussian_integral})} \\
= & (2\pi)^{-\frac{n}{2}} |\vSigma|^{-\frac{1}{2}}  \cos(\vomega\T \vmu) \big|\frac{1}{2}\vSigma^{-1}\big| ^{-\frac{1}{2}}
\pi^{\frac{n}{2}} \exp(-\frac{1}{2} \vomega\T \vSigma \vomega) \\
= & \exp(-\frac{1}{2} \vomega\T \vSigma \vomega)  \cos(\vomega\T \vmu) 
\end{align*}

To prove Eq.~\ref{eqn_x_cos_gaussian_integral}, we first prove that:
\begin{equation} \label{eqn_x_cos_gaussian_integral_single}
\begin{split}
&\int \va\T \vx ~\cos(\vomega\T \vx)  ~p(\vx) ~\rd \vx , \quad \vx \sim \mathcal{N}(\vmu,\vSigma)\\
=& \exp\big(-\frac{\vomega\T\vSigma\vomega}{2}\big)\cos(\vomega\T\vmu) \va\T\vmu 
- \exp\big(-\frac{\vomega\T\vSigma\vomega}{2}\big)\sin(\vomega\T\vmu) \va\T\vSigma\vomega
\end{split}
\end{equation}

\begin{equation}
\begin{split}
&\int \va\T \vx ~\cos(\vomega\T \vx)  ~p(\vx) ~\rd \vx , \quad \vx \sim \mathcal{N}(\vmu,\vSigma)\\
= &\Re \Big\{\int \va\T \vx \exp(i \vomega\T \vx)  (2\pi)^{-\frac{n}{2}} |\vSigma|^{-\frac{1}{2}}
\exp (-\frac{1}{2}(\vx - \vmu)\T \vSigma^{-1} (\vx - \vmu)) \rd \vx
\Big\} \quad \quad\text{(Eq.~\ref{eqn_multivariant_gaussian})} \\
= &(2\pi)^{-\frac{n}{2}} |\vSigma|^{-\frac{1}{2}}  \Re \Big\{\int \big(\va\T \vmu + \va\T(\vx - \vmu)\big)
\exp(i\vomega\T \vmu) \exp(i \vomega\T (\vx-\vmu))  
\exp (-(\vx - \vmu)\T (2\vSigma)^{-1} (\vx - \vmu)) \rd \vx
\Big\} \\
= & \va\T \vmu \exp(-\frac{1}{2} \vomega\T \vSigma \vomega)  \cos(\vomega\T \vmu) 
+  (2\pi)^{-\frac{n}{2}} |\vSigma|^{-\frac{1}{2}} 
\Re \Big\{\exp(i\vomega\T \vmu) \int \va\T \vx  \exp (-\frac{1}{2}\vx \T \vSigma^{-1} \vx  + (i\vomega)\T \vx 
) \rd \vx \\
= & \va\T \vmu \exp(-\frac{1}{2} \vomega\T \vSigma \vomega)  \cos(\vomega\T \vmu) 
+  (2\pi)^{-\frac{n}{2}} |\vSigma|^{-\frac{1}{2}} 
\Re \Big\{\exp(i \vomega\T \vmu)  ~
i\va\T \vSigma \vomega ~
\pi^{\frac{n}{2}} |\frac{1}{2}\vSigma^{-1}|^{-\frac{1}{2}} \exp(\frac{1}{4} (i\vomega)\T 2 \vSigma (i \vomega))
\Big\} \\
= & \va\T \vmu \exp(-\frac{1}{2} \vomega\T \vSigma \vomega)  \cos(\vomega\T \vmu) 
- \va \T \vSigma \vomega \exp(-\frac{1}{2} \vomega\T \vSigma \vomega)  \sin(\vomega\T \vmu) \\
\end{split}
\end{equation}
Then we can select $\va$ as $\ve_i$, which is a vector with only $i$th element $1$, and others $0$. After stacking these elements together, we can get Eq.~\ref{eqn_x_cos_gaussian_integral}.

\subsection{Predictive Distribution with Uncertain Input}
\subsubsection{Predictive Mean}
For the simplicity in notation, we neglect the tilde on top of $\tx$ as the combined state, and treat $\vx$ as the test input $\tx^*$. Furthermore, we define
\begin{equation}
\es(\vx) \defeq \exp(-\frac{\vx\T \vSigma \vx}{2} )\sin(\vx\T \vmu) \quad
\ec(\vx) \defeq \exp(-\frac{\vx\T \vSigma \vx}{2} )\cos(\vx\T \vmu) \quad
\end{equation}
\begin{equation}
\bar i \defeq \text{mod} (i, r)
\end{equation}

The predictive mean with uncertain test input $\vx\sim\mathcal{N}(\vmu,\vSigma) \in\vR$ can be derived using the law of total expectation, and Eq.\ref{eqn_cos_gaussian_integral} and \ref{eqn_sin_gaussian_integral}:
\begin{equation}
\begin{split}
\mu_f
&:= \E[f(\vx)|\vmu,\vSigma] = \E_{\vx}[\E_f[f(\vx)]] \\
&=\E_{\vx}[\vw\T \vphi(\vx)] = \vw\T  \E_{\vx}[\vphi(\vx)] \\ 
&= \vw\T \frac{\sigma_f}{\sqrt{r}} 
\begin{bmatrix}
\int \cos(\VOmega\T\vx) p(\vx) \rd \vx \\
\int \sin(\VOmega\T\vx) p(\vx) \rd \vx
\end{bmatrix} \\
&= \vw \T \vq
\end{split}
\end{equation}
\begin{equation*}
\vq_i := \frac{\sigma_f}{\sqrt{r}} \;
\begin{cases}
\ec(\vomega_{i}) \quad
& i \leq r\\
\es(\vomega_{\bar i}) & i > r
\end{cases}
\end{equation*}

\subsubsection{Predictive Variance}
Next we compute the predictive variance using the law of total variance, and Eq.~\ref{eqn_cos_cos_gaussian_integral}, \ref{eqn_sin_sin_gaussian_integral}, and \ref{eqn_cos_sin_gaussian_integral}:
\begin{equation}
\begin{split}
\Sigma_f &:= \Var[f(\vx)|\vmu,\vSigma] \\
&= \E_{\vx}[\Var_f[f(\vx)]] + \Var_{\vx}[\E_f[f(\vx)]] \\
&= \E_{\vx}[\Var_f[f(\vx)]] + \E_{\vx}[\E_f[f(\vx)]^2] - \E_{\vx}[\E_f[f(\vx)]]^2 \\
&=  \E_{\vx}[\sigma_n^2 (1+ \vphi(\vx)\T\vA^{-1}\vphi(\vx))] + \E_{\vx}[(\vw\T\vphi(\vx))^2 ] - \mu_f^2 \\
&= \sigma_n^2 + \sigma_n^2  \Big (\int \vphi(\vx)\T \vA^{-1} \vphi(\vx) p(\vx) \rd \vx \Big)
+ \vw\T \Big(\int \vphi(\vx) \vphi(\vx)\T p(\vx) \rd \vx\Big) \vw - \mu_f^2 \\
&= \sigma_n^2 + \Tr \Big( 
\underbrace {\big(\sigma_n^2 \vA^{-1} + \vw \vw\T \big)}_\vS
\underbrace {\int \vphi(\vx) \vphi(\vx)\T p(\vx) \rd \vx}_\vT
\Big) - \mu_f^2
\end{split}
\end{equation}
\begin{align}
\vT_{ij} :=&
\frac{\sigma_f^2}{2r}  \cdot \;
\begin{cases}
+\ec(\vtheta_1) + \ec(\vtheta_2) \quad&
i\leq r , j \leq r
\\
+\es(\vtheta_1) - \es(\vtheta_2)&
i\leq r, j > r
\\
+\es(\vtheta_1) + \es(\vtheta_2)&
i > r, j \leq r
\\
-\ec(\vtheta_1) + \ec(\vtheta_2)&
i > r, j > r
\end{cases}
\\
\vtheta_1  :=& \vomega_{\bar i} + \vomega_{\bar j} \quad \vtheta_2 := \vomega_{\bar i} - \vomega_{\bar j}
\end{align}

\subsubsection{Covariance between Input and Prediction}
The covariance between input and prediction can be computed as:
\begin{equation}
\begin{split}
\vSigma_{\vx,f} &:= \Cov [\vx, f(\vx) | \vmu, \vSigma] 
\approx \Cov_{\vx}[\vx, \E_f[f(\vx)]] 
=  \Cov_{\vx}[\vx, \vw\T \vphi(\vx)]  \\
&=  \E_{\vx}[(\vx-\vmu) (\vw\T \vphi(\vx) - \vw\T \vq)] \\
&=  \E_{\vx}[\vw \T \vphi(\vx) \vx - \vw \T \vq \vx - \vw\T \vphi(\vx) \vmu + \vw\T \vq \vmu ] \\ 
&=  \Big(\sum_{i=1}^{2r} \vw_i \underbrace{\E_{\vx}[\vphi_i(\vx) \vx]}_{\vP_i} \Big)
- \vw\T \vq \vmu 
- \vw\T \E_{\vx}[\vphi(\vx)]  \vmu + \vw\T \vq \vmu \\ 
&=  \vP  \vw  - \vw \T \vq \vmu
\end{split}
\end{equation}

\begin{equation}
\vP_i := \frac{\sigma_f}{\sqrt{r}}  \cdot
\begin{cases}
\ec(\vomega_i)\vmu - \es(\vomega_i) \vSigma\vomega_i \quad
& i \leq r \\
\es(\vomega_{\bar i}) \vmu + \ec(\vomega_{\bar i}) \vSigma\vomega_{\bar i}
 & i > r
\end{cases}
\end{equation}

\subsubsection{Covariance between Two Predictive Outputs}
Here we compute the covariance of two different outputs $f^i$, $f^j$ with uncertain input.
\begin{equation}
\begin{split}
\Cov[f^i, f^j |\vmu, \vSigma] &= \E_\vx [\E[f^i] \E[f^j]] -  
\E_\vx [\E[f^i]] \E_\vx [\E[f^j]] \\
&= \E_\vx [ ({\vw^i}\T \vphi^i) ( {\vw^j}\T \vphi^j)  ] -  \mu_f^i \mu_f^j \\
&= {\vw^i}\T  \E_\vx [ \vphi^i  {\vphi^j}\T  ]  {\vw^j}-  \mu_f^i \mu_f^j \\
&= {\vw^i}\T 
\underbrace{
\Big( \int \vphi^i {\vphi^j}\T p(\vx) \rd \vx\ \Big) }
_{\vT}
{\vw^j} -  \mu_f^i \mu_f^j \\
\end{split}
\end{equation}

\begin{align}
\vT_{st} &:=
\frac{\sigma_f^2}{2r}  \cdot \;
\begin{cases}
+\ec(\vtheta_1) + \ec(\vtheta_2) \quad&
s\leq r , t \leq r
\\
+\es(\vtheta_1) - \es(\vtheta_2)&
s \leq r, t > r
\\
+\es(\vtheta_1) + \es(\vtheta_2)&
s > r, t \leq r
\\
-\ec(\vtheta_1) + \ec(\vtheta_2)&
s > r, t > r
\end{cases}
\\
\vtheta_1  &:= \vomega^i_{\bar s} + \vomega^j_{\bar t} \quad 
\vtheta_2 := \vomega^i_{\bar s} - \vomega^j_{\bar t}
\end{align}
where superscript $i$ denotes the corresponding value related to the $i$th output, e.g. $\vw^i$, $\vphi^i$, and $\mu_f^i$ are the coefficients, features mapping, and predictive mean of the $i$th output, respectively.

\subsection{Derivative of Predictive Distribution to Uncertain Input}
In this document, we use denominator layout for matrix calculus.
\subsubsection{Predictive Mean}

\begin{equation}
\begin{split}
\frac{\partial \mu_f} {\partial \vmu}
= \frac{\partial \wb \T \qb} {\partial \vmu} 
= \frac {\partial \qb} {\partial \vmu} \wb 
\end{split}
\end{equation}

\begin{equation}
\frac{\partial \qb_i} {\partial \vmu} =  
\frac{\sigma_f}{\sqrt{r}} \cdot
\begin{cases}
-\es(\vomega_i) \vomega_i \quad
& i \leq r \\
+\ec(\vomega_{\bar i}) \vomega_{\bar i} & i > r
\end{cases}
\end{equation}

\begin{equation}
\frac{\partial \mu_f} {\partial \vSigma} =
\frac{\partial \wb \T \qb} {\partial \vSigma} =
\sum_i \vw_i
\frac{\partial \qb_i} {\partial \vSigma} =
\frac {\partial \qb} {\partial \vSigma} \wb  
\end{equation}

\begin{equation}
\frac {\partial \qb_i} {\partial \vSigma} = 
-\frac{\sigma_f}{2 \sqrt{r}}  \cdot
\begin{cases}
 \ec(\vomega_i) \vomega_{i} \vomega_{i}\T \quad&
i \leq r \\
\es(\vomega_{\bar i}) \vomega_{\bar i} \vomega_{\bar i} \T &
i > r 
\end{cases}
\end{equation}

\subsubsection{Predictive Covariance}
\begin{equation}
\begin{split}
\frac{\partial \Sigma_f} {\partial \vmu} =& 
\frac{\partial \big(\sigma_n^2 + \Tr(\vS \vT) - \mu_f^2 \big)} {\partial \vmu} \\
= &
\frac{\partial \Tr(\vS \vT)} {\partial \vmu} -
\frac{\partial \mu_f^2} {\partial \vmu} \\
= &
\sum_{i,j}\vS_{ij} \frac{\partial \vT_{ij}}{\partial \vmu} - 
2\frac {\partial \mu_f} {\partial \vmu} \mu_f \\
\end{split}
\end{equation}

\begin{equation}
\begin{split}
\frac{\partial \vT_{ij}}  {\partial\vmu} &=
\frac{\sigma_f^2}{2r}  \cdot \;
\begin{cases}
-\es(\vtheta_1)\vtheta_1 - \es(\vtheta_2)\vtheta_2 &
i\leq r , j \leq r
\\
+\ec(\vtheta_1)\vtheta_1 - \ec(\vtheta_2)\vtheta_2&
i\leq r, j > r
\\
+\ec(\vtheta_1) \vtheta_1 + \ec(\vtheta_2)\vtheta_2&
i > r, j \leq r
\\
+\es(\vtheta_1)\vtheta_1 - \es(\vtheta_2)\vtheta_2&
i > r, j > r
\end{cases}
\\
\vtheta_1 &:= \vomega_{\bar i} + \vomega_{\bar j} 
\quad \vtheta_2 := \vomega_{\bar i} - \vomega_{\bar j}
\end{split}
\end{equation}

\begin{equation}
\begin{split}
\frac{\partial \Sigma_f} {\partial \vSigma} &=
\sum_{i,j}\vS_{ij} \frac{\partial \vT_{ij}} {\partial \vSigma} - 
2\frac {\partial \mu_f} {\partial \vSigma} \mu_f \\
\end{split}
\end{equation}

\begin{align}
\frac{\partial \vT_{ij}} {\partial \vSigma} &=
-\frac{\sigma_f^2}{4r}  \cdot \;
\begin{cases}
+\ec(\vtheta_1)\vtheta_1 \vtheta_1\T + \ec(\vtheta_2) \vtheta_2 \vtheta_2\T \quad&
i\leq r , j \leq r
\\
+\es(\vtheta_1) \vtheta_1 \vtheta_1 \T - \es(\vtheta_2) \vtheta_2 \vtheta_2 \T&
i\leq r, j > r
\\
+\es(\vtheta_1) \vtheta_1 \vtheta_1 \T + \es(\vtheta_2)\vtheta_2 \vtheta_2 \T&
i > r, j \leq r
\\
-\ec(\vtheta_1) \vtheta_1 \vtheta_1 \T+ \ec(\vtheta_2) \vtheta_2 \vtheta_2 \T&
i > r, j > r
\end{cases}
\\
\vtheta_1  &:= \vomega_{\bar i} + \vomega_{\bar j} \quad 
\vtheta_2 := \vomega_{\bar i} - \vomega_{\bar j}
\end{align}

\subsubsection{Covariance between Input and Prediction}

\begin{equation}
\begin{split}
\frac{\partial \vSigma_{\vx,f} } {\partial \vmu} &=
\Big(\sum_{i=1}^{2r} \vw_i \frac{\partial  \vP_i}{\partial \vmu} \Big)- 
\frac{\partial \vw\T\vq \vmu} {\partial \vmu}
\\
&= \Big(\sum_{i=1}^{2r} \vw_i \frac{\partial  \vP_i}{\partial \vmu} \Big) - 
\Big(
\frac{\partial \mu_f} {\partial \vmu} \vmu\T + \mu_f \Ib
\Big)
\end{split}
\end{equation}

\begin{equation}
\frac{\partial \vP_i} {\partial \vmu} = \frac{\sigma_f}{\sqrt{r}} \;
\begin{cases}
\ec(\vomega_{i})\Ib - \es(\vomega_i) \vomega_i \vmu \T - \ec(\vomega_i) \vomega_i  \vomega_i\T  \vSigma \quad
& i \leq r\\
\es(\vomega_{\bar i}) \Ib + \ec(\vomega_{\bar i}) \vomega_{\bar i} \vmu\T -
\es(\vomega_{\bar i}) \vomega_{\bar i}  \vomega_{\bar i}\T \vSigma
 & i > r
\end{cases}
\end{equation}

\begin{equation}
\begin{split}
\frac{\partial \vSigma_{\vx,f} } {\partial \vSigma} &=
\Big(\sum_{i=1}^{2r} \vw_i \frac{\partial  \vP_i}{\partial \vSigma} \Big)- 
\frac{\partial \mu_f \vmu} {\partial \vSigma}
\\
\end{split}
\end{equation}

\begin{equation}
\frac{\partial \vP_{ij}} {\partial \vSigma} = -\frac{\sigma_f}{2\sqrt{r}} \;
\begin{cases}
\ec(\vomega_i) \vomega_i \vomega_i\T \vmu_j - \es(\vomega_i) \vomega_i \vomega_i\T (\vSigma \vomega_i)_j
& i \leq r\\
\es(\vomega_{\bar i}) \vomega_{\bar i} \vomega_{\bar i} \T \vmu_j + \ec(\vomega_{\bar i}) \vomega_{\bar i} \vomega_{\bar i}\T (\vSigma \vomega_{\bar i})_j
 & i > r
\end{cases}
\end{equation}

\begin{equation}
\frac{\partial \mu_f \vmu_i} {\partial \vSigma} = 
\vmu_i \frac{\partial \mu_f } {\partial \vSigma} 
\end{equation}


\section{Detailed Derivation of Linearization Methods}

We derive the prediction distribution and its derivatives in I-SSGPR with uncertain input. In particular, the predictive distribution is achieved through approximated using the first-order Taylor expansion of the predictive mean. We use the denominator-layout notation for matrix calculus, in which case the derivative of a scalar to a vector is a column vector. For the simplicity in notation, we neglect the tilde on top of $\tx$ as the combined state.

We define
\begin{equation}
\bar i := \text{mod} (i, r)
\end{equation}

The input is uncertain, and is assumed to have a normal distribution:
\begin{equation}
\vx \sim \mathcal{N} (\vmu, \vSigma)
\end{equation}

The predictive distribution of I-SSGPR with certain input:
\begin{equation}
f(\vx) |\vx \sim \mathcal{N} (\vw \T \vphi(\vx), 
\sigma_n^2  \big( 1 + \vphi(\vx)\T \vA^{-1} \vphi(\vx) \big)
\end{equation}
\begin{equation}
\vphi(\vx) := 
\begin{bmatrix} \vc(\vx) \\ \vs(\vx) \end{bmatrix},
\quad \vc(\vx) := \frac{\sigma_f}{\sqrt{r}} \cos(\VOmega\T\vx)
\quad \vs(\vx) := \frac{\sigma_f}{\sqrt{r}} \sin(\VOmega\T\vx)
\end{equation}
where $\vphi(\vx)$ is the feaure mapping.
The partial derivative of the predictive mean $\E_f[f(\vx)]$ to the input $\vx$ is:
\begin{equation}
\frac{\partial \E_f[f(\vx)]}{\partial \vx} = 
\frac{\partial \big( \vw \T \vphi(\vx) \big)} {\partial \vx} =
\frac{\partial\vphi(\vx)} {\partial \vx} \vw =
\vD(\vx) \vw
\end{equation}
\begin{equation}
\vD_i(\vx) := 
\frac{\sigma_f}{\sqrt{r}} \cdot
\begin{cases}
-\vomega_i \sin(\vomega_i\T\vx) \quad
& i \leq r\\
+\vomega_{\bar i} \cos(\vomega_{\bar i}\T\vx) \quad
& i > r
\end{cases}
\end{equation}
where $\vD_i(\vx)$ is the $i$th column of $\vD(\vx)$.

\subsection{Predictive Distribution with Uncertain Input}
\subsubsection{Predictive Mean}
We obtain the predictive mean with uncertain input by computing the predictive GP mean at input mean $\vmu$. Therefore 
\begin{equation}
\begin{split}
\mu_f &:= \E[f(\vx)|\vmu,\vSigma] =  \E_{f}[ \E_{\vx}[f(\vx)] ] \\
&\approx \E_f[f(\vmu)|\vmu] = \vw\T\vphi(\vmu)
\end{split}
\end{equation}

\subsubsection{Predictive Variance}
We first linearize the predictive mean, and then compute the predictive variance with uncertain input based on the law of total variance. 
\begin{equation}
\begin{split}
\E_f[f(\vx)] \approx& \E_f[f(\vmu)] + \frac{\partial \E_f[f(\vx)]}{\partial \vx}\bigg|^{\rT}_{\vx=\vmu}\Big(\vx - \vmu\Big) \\
=&\vw\T \vphi(\vmu) + (\vD(\vmu)\vw)\T (\vx- \vmu) \\
=& \underbrace{ (\vD(\vmu)\vw)\T}_{\va(\vmu)\T} \vx + \underbrace{\vw\T \vphi(\vmu) - (\vD(\vmu)\vw)\T \vmu}_{b(\vmu)}
\end{split}
\end{equation}

To simplify notation, from now on, we denote $\vD = \vD(\vmu)$, $\va$ = $\va(\vmu)$, and $b = b(\vmu)$.

\begin{equation}
\begin{split}
\Sigma_f &:= \Var [f(\vx)|\vmu,\vSigma]  \\
&= \Var_\vx [\E_f [f(\vx)]  ] + \E_\vx [\Var_f [f(\vx)] ] \\
& \approx \Var_\vx [\va\T \vx + b ] + \Var_f[f(\vmu)] \\
&= \Var_\vx [\va\T \vx] + \Var_f[f(\vmu)] \\
&=\va\T \Var_\vx [\vx ]\;\va + \Var_f[f(\vmu)] \\
&= \va\T \vSigma \;\va + \sigma_n^2\Big(1+ \vphi(\vmu)\T\vA^{-1}\vphi(\vmu)\Big) 
\end{split}
\end{equation}

\subsubsection{Covariance between Input and Prediction}
The covariance between input and prediction can be computed as:

\begin{equation}
\begin{split}
\vSigma_{\vx,f} := &\Cov[\vx, f(\vx) | \vmu, \vSigma]  \\
\approx &\Cov_{\vx}[\vx, \E_f[f(\vx)]] 
\\
= & \Cov_{\vx}[\vx, \va\T \vx + b]  \\
= & \E_{\vx}[(\vx - \vmu) (\va\T \vx - \va\T \vmu)]  \\
= & \E_{\vx}[(\vx - \vmu) (\vx - \vmu)\T \va]  \\ 
= & \vSigma \va
\end{split}
\end{equation}

\subsubsection{Covariance between Two Predictive Outputs}
\begin{equation}
\begin{split}
\Cov[f^i, f^j |\vmu, \vSigma] 
&= \E_\vx \Big[ \Big(\E_f[f^i] - \E_\vx[\E_f[f^i]]\Big)  \Big(\E_f[f^j] - \E_\vx[\E_f[f^j]]\Big)\Big] \\
&= \E_\vx [{\va^i}\T (\vx-\vmu) \;{\va^j}\T(\vx-\vmu)] \\
&={\va^i}\T \vSigma \va^j
\end{split}
\end{equation}
where we use subscript $i$ to denote the corresponding value for the $i$th output.

\subsection{Derivative of Predictive Distribution to Uncertain Input}
\subsubsection{Derivative of Predctive Mean}
Derivative of predictive mean with respect to the input mean:
\begin{equation}
\frac{\partial \mu_f }{\partial \vmu} = \frac{\partial \big( \vw\T\vphi(\vmu)\big)}{\partial \vmu} = \vD \vw
\end{equation}

Derivative of predictive mean with respect to the input variance:
\begin{equation}
\frac{\partial \mu_f }{\partial \vSigma} = \bf 0
\end{equation}

\subsubsection{Derivative of Predictive Variance}

Derivative of predictive variance with respect to the input mean:
\begin{equation}
\begin{split}
\frac{\partial \Sigma_f}{\partial \vmu} &= 
\frac{\partial \Big(\va\T \vSigma \va + \sigma_n^2 \big( 1 + \vphi(\vmu)\T \vA^{-1} \vphi(\vmu)\big)
\Big)}{\partial \vmu}\\
&= \frac{ \partial \big(\va\T \vSigma \va \big)} {\partial \vmu}
+ \sigma_n^2 \frac{\big( \vphi(\vmu)\T \vA^{-1} \vphi(\vmu) \big)} {\partial \vmu} \\
&= 2 \frac{\partial \big(\vD \vw \big)}{\partial \vmu} \vSigma \vD \vw
+ 2\sigma_n^2 \frac{\partial \vphi(\vmu)}{\partial \vmu} \vA^{-1} \vphi(\vmu) \\
&= 2  \Big(\sum_{i=1}^{2r} \vw_i \frac{\partial \vD_i}{\partial \vmu} \Big)\vSigma \vD \vw
+ 2\sigma_n^2  \vD \vA^{-1} \vphi(\vmu) \\
\end{split}
\end{equation}

\begin{equation}
\frac{\partial \vD_i}{\partial \vmu} =
\frac{\sigma_f}{\sqrt{r}} \cdot
\begin{cases}
-\vomega_i \cos(\vomega_i\T\vmu)\vomega_i\vomega_i\T \quad
& i \leq r\\
-\vomega_{\bar i} \sin(\vomega_{\bar i}\T\vmu) \vomega_{\bar i}\vomega_{\bar i}\T\quad
& i > r\\
\end{cases}
\end{equation}



Derivative of predictive variance with respect to the input variance:
\begin{equation} \frac{\partial \Sigma_f} {\partial \vSigma} =
\frac{\partial \big( \va\T \vSigma \va \big) }{\partial
\vSigma} = \va \va\T
\end{equation}

\subsubsection{Covariance between Input and Prediction}
Its derivatives to the input mean and variance are:
\begin{equation}
\begin{split}
\frac{\partial \vSigma_{\vx,f}} {\partial \vmu} = 
\frac{\partial \vSigma \va } {\partial \vmu} = 
\frac{\partial \va } {\partial \vmu} \vSigma = \vV \vSigma
\end{split}
\end{equation}

\begin{equation}
\begin{split}
\frac{\partial \vSigma_{\vx,f}} {\partial \vSigma} = 
\frac{\partial \vSigma  \va} {\partial \vSigma} = \vT, \quad
\vT_{ijk} = \va_j, \text{if} \;\;i = k,  \;\text{otherwise}\;\; \vT_{ijk} = 0
\end{split}
\end{equation}

\section{Differential Dynamic Programming in Belief Space}
The Bellman equation  for the value function  in discrete-time is specified  as follows
 \begin{align}
V(\vb_k,k) = \min_{\vu_k}\Bigg[\underbrace{\mathcal{L}(\vb_k,\vu_k) + V\Big(\mF(\vb_k,\vu_k),k+1 \Big)}_{Q(\vb_k,\vu_k)}\Bigg].
\end{align}
The cost $\mathcal{L}(\vb_k,\vu_k)$ is the expectation of a quadratic cost function \small
\begin{equation}\label{cost_exp}
\begin{split}
&\mathcal{L} (\vb_k,\vu_k) = \mathbb{E}_{\vx}\Big[\mathcal{L} (\vx_k,\vu_k)\Big] = \rtr(\vSigma_k\vQ) + (\vmu_k-\vx_k^{goal})^{\rT}\vQ (\vmu_k-\vx_k^{goal})
 + \vu_k^{\rT}\vR\vu_k. 
\end{split}
\end{equation} \normalsize
We create a quadratic local model of the value function by expanding the $Q$-function up to the second order
\begin{equation}
\begin{split}
Q_k(\vb_k+\delta \vb_k,\vu_k+\delta \vu_k)\approx Q_k^0+Q^\vb_k\delta\vb_k+Q^\vu_k\delta\vu_k +\frac{1}{2}\left[ \begin{array}{c} \delta\vb_k \\ \delta\vu_k \end{array}\right]^{\rT} \left[\begin{array}{cc}Q^{\vb \vb}_k & Q^{\vb \vu}_k\\ Q^{\vu \vb}_k & Q^{\vu \vu}_k \end{array}\right] \left[ \begin{array}{c} \delta\vb_k \\ \delta\vu_k \end{array}\right],
\end{split}
\end{equation}
where the superscripts of the $Q$-function indicate derivatives. For instance, $Q^b_k=\nabla_bQ_k(\vb_k,\vu_k)$. We use this superscript rule for $\mathcal{L}$ and $V$ as well. To find the optimal control policy, we compute the local variations in control $\delta\hat{\vu}_k$ that maximize the $Q$-function
\begin{equation}\label{policy}
\begin{split}
\delta\hat{\vu}_k &= \arg\min_{\delta\vu_k} \Big[Q_k(\vb_k+\delta \vb_k,\vu_k+\delta \vu_k) \Big] \\
&= \underbrace{-(Q^{\vu \vu}_k)^{-1}Q_k^\vu}_{\vI_k}  \underbrace{-(Q^{\vu \vu}_k)^{-1}Q_k^{\vu \vb}}_{\vL_k} \delta\vb_k .
\end{split}
\end{equation}
The optimal control can be found as $\hat{\vu}_k=\bar{\vu}_k+\delta\hat{\vu}_k$.  The control policy is a linear function of the belief $\vb_k$, therefore the controller is deterministic. 
The quadratic expansion of the value  function is backward propagated  based on the equations that follow
\begin{align}\label{back}
Q^{\vb}_k&=\mathcal{L}_k^z + V_k^b\mF^b_k,~~~Q^{\vu}_k=\mathcal{L}_k^u + V_k^b\mF^u_k,\nonumber\\
Q^{\vb \vb}_k&=\mathcal{L}_k^{\vb \vb} +(\mF^\vb_k)^{\rT}V^{\vb \vb}_k \mF^\vb_k,~~~Q^{\vu \vb}_k=\mathcal{L}_k^{\vu \vb} +(\mF^\vu_k)^{\rT}V^{\vb \vb}_k \mF^\vb_k, \nonumber\\
Q^{\vu \vu}_k&=\mathcal{L}_k^{\vu \vu} +(\mF^\vu_k)^{\rT}V^{\vv \vv}_k \mF^\vu_k, ~~~~V_{k-1}=V_k+Q^\vu_k\vI_k \nonumber\\
V^\vv_{k-1} &= Q^{\vb}_k + Q^\vu_k\vL_k,~~~~~~ V^{\vv \vv}_{k-1} = Q^{\vb \vb}_k + Q^{\vb \vu}_k\vL_k   . 
\end{align}

Next we provide the DDP algorithm in belief space  in pseudocode form. 
\begin{algorithm}
\caption{DDP in belief space }\label{algorithm_pddp}
\begin{algorithmic}[1] 
\State \textbf{Initialization:}  Given the nominal trajectory $(\bar{\vb}_k,\bar{\vu}_k)$
\Repeat
\State\parbox[t]{\dimexpr\linewidth-\algorithmicindent-\algorithmicindent\relax}{\textbf{Local approximation:} Obtain linear approximation of the belief dynamics along a nominal trajectory $(\bar{\vb}_k,\bar{\vu}_k)$. \strut}
\State\parbox[t]{\dimexpr\linewidth-\algorithmicindent-\algorithmicindent\relax}{\textbf{Backward sweep:} Compute the quadratic approximation of the value function (\ref{back}) and obtain optimal policy for control correction $\delta\hat{\vu}_k=\vI_k+\vL_k \delta\vb_k$ (\ref{policy}).\strut}
\State\parbox[t]{\dimexpr\linewidth-\algorithmicindent-\algorithmicindent\relax}{\textbf{Forward sweep:} Update control $\bar{\vu}_k=\bar{\vu}_k+\delta\hat{\vu}_k$ and perform approximate inference to obtain a new nominal trajectory $(\bar{\vb}_k,\bar{\vu}_k)$. \strut}
\Until Termination condition is satisfied  \\
\Return Optimal state and control trajectory. 
\end{algorithmic}
\end{algorithm}

\subsection{Control constraint}\label{Control_constraints}
 Control constrains can be taken into account in different fashions, it has been shown that using naive clamping and squashing functions performs unfavorably than directly incorporating the constraints while minimizing the $Q$-function \cite{tassacontrol}. In this work we take into account the control constraints by solving a quadratic programming (QP) problem subject to a box constraint
\begin{equation}\label{QP}\small
\begin{split}
\min_{\delta\vu_k} ~~Q_k(\vb_k+\delta \vb_k,\vu_k+\delta \vu_k) ~~\text{s. t}  ~~\vu_{\min}\leq \vu_k + \delta\vu_k  \leq \vu_{\max}
\end{split}
\end{equation}\normalsize
where $\vu_{\min}$ and $\vu_{\max}$ correspond to the lower and upper bounds of the controller.
The QP problem (\ref{QP}) can be solved efficiently due to the fact that at each time step  the scale of the QP problem is relatively small. And warm-start can be used to further speed up the optimization in the backward pass. Solving (\ref{QP}) directly is not feasible since $\delta\vb$ is not known in the backward sweep. The optimum consists of feedforward and feedback parts, i.e., $\vI_k+\vL_k \delta\vb_k=\arg\min_{\delta\vu_k} \big[Q_k(\vb_k+\delta \vb_k,\vu_k+\delta \vu_k) \big]$, here we adopt the strategy in  \cite{tassacontrol} using the Projected-Newton algorithm \cite{bertsekas1982projected}. The feedforward gain is computed by solving the QP problem
\begin{equation}\label{QP_feedforward}\small
\begin{split}
\vI_k = \arg\min_{\delta\vu_k} \Big[\delta\vu_k\T Q^{uu}_k \delta\vu_k + Q^{b}_k\delta\vu_k \Big]~~\text{s. t.}  ~~\vu_{\min}\leq \vu_k + \delta\vu_k  \leq \vu_{\max}
\end{split}
\end{equation}\normalsize
The algorithm  gives the decomposition of \scriptsize $Q^{uu}_k=\left[ \begin{array}{cc} Q^{uu}_{k,ff} & Q^{uu}_{k,fc}\\Q^{uu}_{k,cf} & Q^{uu}_{k,cc} \end{array} \right]$\normalsize where the indices $f,c$ correspond to clamped (when $\vu_k=\vu_{\min}$ or $\vu_{\max}$) of free ($\vu_{\min}<\vu_k<\vu_{\max}$) parts, respectively. . The feedback gain associated to the free part is obtained by $\vL_k=-Q^{uu}_{k,ff}Q^{ub}_k$. The rows of $\vL_k$ corresponding to clamped controls are set to be zero.



\end{document}